# DON'T KILL THE BABY!
## THE CASE FOR AI IN ARBITRATION

### Michael J. Broyde* & Yiyang Mei**

*Since the introduction of Generative AI (GenAI) in 2022, its ability to simulate human intelligence and generate content has sparked both enthusiasm and concern. While much of the criticism focuses on AI's potential to perpetuate bias, create emotional dissonance, displace jobs, and raise ethical questions, these concerns often overlook the practical benefits of AI, particularly in legal contexts. This article examines the integration of AI into arbitration, arguing that the Federal Arbitration Act (FAA) allows parties to contractually choose AI-driven arbitration, despite traditional reservations.*

*This article makes three key contributions: (1) It shifts the focus from debates over AI's personhood to the practical aspects of incorporating AI into arbitration, asserting that AI can effectively serve as an arbitrator if both parties agree; (2) It positions arbitration as an ideal starting point for broader AI adoption in the legal field, given its flexibility and the autonomy it grants parties to define their standards of fairness; and (3) It outlines future research directions, emphasizing the importance of empirically comparing AI and human arbitration, which could lead to the development of distinct systems.*

*By advocating for the use of AI in arbitration, this article underscores the importance of respecting contractual autonomy and creating an environment that allows AI's potential to be fully realized. Drawing on the insights of Judge Richard Posner, the article argues that the ethical obligations of AI in arbitration should be understood within the context of its technological strengths and the voluntary nature of arbitration agreements. Ultimately, it calls for a balanced, open-minded approach to AI in arbitration, recognizing its potential to enhance the efficiency, fairness, and flexibility of dispute resolution.*

---

\* Michael J. Broyde is a Professor of Law at Emory University, the Berman Projects Director in its Center for the Study of Law and Religion, and the Director of the SJD Program at Emory University School of Law.

\*\* Yiyang Mei is an SJD candidate at Emory University, where she focuses on the intersection of law and technology. She holds a JD and an MPH in epidemiology, also from Emory.

INTRODUCTION

Since the introduction of Generative AI (GenAI) in 2022, extensive discussions have highlighted its impressive capabilities in simulating human intelligence and generating novel content.[1] Despite these remarkable functionalities, there is consensus that, if not carefully managed and fine-tuned, GenAI—especially when applied through Large Language Models (LLMs)[2]—can be fundamentally harmful and discriminatory toward marginalized groups.[3] In the workplace, Artificial

---

1. *See* Cole Stryker & Eda Kavlakoglu, *What is Artificial Intelligence (AI)?*, IBM, https://www.ibm.com/topics/artificial-intelligence (last visited Aug. 3, 2024) (AI is a technology that enables computers and machines to simulate human intelligence and problem-solving capabilities); John Nosta, *When Artificial Intelligence Mimics the Human Brain*, PSYCH. TODAY (Nov. 30, 2023), https://www.psychologytoday.com/us/blog/the-digital-self/202311/when-artificial-intelligence-mimics-the-human-brain; Aditya Roy, *Simulating Human Intelligence: Are We Close to It?*, MEDIUM (July 5, 2021), https://becominghuman.ai/simulating-human-intelligence-are-we-close-to-it-463976a4c203; Pandora Dewan, *New AI Can Mimic Human Brain, Use the Same Tricks as We Do*, NEWSWEEK (Nov. 20, 2023, 11:00 AM), https://www.newsweek.com/new-ai-mimic-human-brain-tricks-neuroscience-1845159; Ana Santos Rutschman, *Artificial Intelligence Can Now Emulate Human Behaviors – Soon It Will Be Dangerously Good*, PHYS ORG (Apr. 5, 2019), https://phys.org/news/2019-04-artificial-intelligence-emulate-human-behaviors.html; *see also* Jen Hunsaker, *12 AI Content Generators to Make Great Content in Minutes*, SEMRUSH BLOG (July 22, 2024), https://www.semrush.com/blog/ai-content-generators/ (suggesting that AI can create a month's worth of content for one's website, social media, and ad campaigns in a few hours); Thomas H. Davenport & Nitin Mittal, *How Generative AI is Changing Creative Work*, HARV. BUS. REV. (Nov. 14, 2022), https://hbr.org/2022/11/how-generative-ai-is-changing-creative-work; Mihaela Bidilică, *How to Use AI to Write a Book. Overcome Writer's Block with* AI *Assistance*, PUBLISHDRIVE (Apr. 30, 2024), https://publishdrive.com/how-to-use-ai-to-write-a-book.html (recommending that AI writing tools help users generate text using concepts like natural language generation and machine learning).

2. *See What are LLMs?*, IBM, https://www.ibm.com/topics/large-language-models (last visited Aug. 4, 2024).

3. *See* Olga Akselrod, *How Artificial Intelligence Can Deepen Racial and Economic Inequalities*, ACLU (July 13, 2021), https://www.aclu.org/news/privacy-technology/how-artificial-intelligence-can-deepen-racial-and-economic-inequities (suggesting that there's ample evidence of the discriminatory harm that AI tools can cause to already marginalized groups, as AI is built by humans and deployed in systems and institutions that have been marked by entrenched discrimination – from the criminal legal system, to housing, to the workplace, to our financial system); *see also* Khari Johnson, *A Move for "Algorithmic Reparation" Call for Racial Justice in AI*, WIRED (Dec. 23, 2021, 7:00 AM), https://www.wired.com/story/move-algorithmic-reparation-calls-racial-justice-ai/(explaining how algorithms used to screen apartment renters and mortgage applicants disproportionately disadvantaged Black people due to the historical patterns of segregation that have poisoned the data on which Intelligence (AI) systems used for hiring, promotions, and other Human Resources (HR) tasks may perpetuate biases by learning from historical data.[4] Additionally, AI chatbots and virtual assistants, though capable of mimicking human interactions, lack genuine empathy and human connection.[5] This can lead to emotional dissonance, where users feel temporarily understood without receiving the deeper emotional support that real human relationships provide.[6] Furthermore, AI's efficiency in automating tasks could result in significant job displacement,[7] and its proficiency in analyzing large datasets might enable the creation of personalized content, raising critical ethical questions about consent, manipulation, privacy, and the potential impact on democracy and personal autonomy.[8]

_____________

many algorithms are built); Elisa Jillson, *Aiming for Truth, Fairness, and Equity in Your Company's Use of AI*, Fed. Trade Comm'n (Apr. 19, 2021), https://www.ftc.gov/business-guidance/blog/2021/04/aiming-truth-fairness-equity-your-companys-use-ai (saying that one consequence of this bias in healthcare is that technological advancements that were meant to benefit all patients may have worsened the healthcare disparities for people of color and as a result, the economic and racial division in our country will only deepen).

4. *See* Lena Kempe, *Navigating the AI Employment Bias Maze: Legal Compliance Guidelines and Strategies*, A.B.A. (Apr. 2024), https://www.americanbar.org/groups/business_law/resources/business-law-today/2024-april/navigating-ai-employment-bias-maze/ (suggesting that AI systems may produce biased results because of limitations in their training data or errors in their programming, with significant legal risks in the hiring and HR contexts); *see also* Maria Diaz, *4 Ways AI is Contributing to Bias in the Workplace*, ZDNET (Mar. 18, 2024, 3:40 AM), https://www.zdnet.com/article/4-ways-ai-is-contributing-to-bias-in-the-workplace/.

5. *See AI Chatbots have Shown They Have an "Empathy Gap" that Children are Likely to Miss*, ScienceDaily (July 2024), https://www.sciencedaily.com/releases/2024/07/240710195430.htm (a study at the University of Cambridge shows that "AI chatbots have frequently shown signs of 'empathy gap' that puts young users at risks of distress or harm, raising the urgent need for 'child-safe AI'"); *see also* Drew Turney, *AI Can "Fake" Empathy but Also Encourage Nazism, Disturbing Study Suggests*, LiveScience (May 29, 2024), https://www.livescience.com/technology/artificial-intelligence/ai-can-fake-empathy-but-also-encourage-nazism-disturbing-study-suggests.

6. *See generally* Lennart Seitz, *Artificial Empathy in HealthCare Chatbots: Does it Feel Authentic?*, Computs. in Hum. Behav.: Artificial Hums., Jan.-July 2024.

7. *See* Rakesh Kochhar, *Which U.S. Workers Are More Exposed to AI on Their Jobs?*, Pew Rsch. Ctr. (July 26, 2023), https://www.pewresearch.org/social-trends/2023/07/26/which-u-s-workers-are-more-exposed-to-ai-on-their-jobs/ (analyzing the types and percentages of jobs that are the most exposed to AI risks).

8. *See* Erik Hermann, *Artificial Intelligence and Mass Personalization of Communication Content – An Ethical and Literacy Perspective*, 24(5) New Media & Soc'y 1258, 1259 (2021); *see also* Marietjie Botes, *Autonomy and the Social*

While criticisms of GenAI are valid and well-founded, they often overlook the technology's efficiency and cost-effectiveness. GenAI excels in processing and analyzing data at speeds unattainable by humans,[9] reducing overall costs,[10] accelerating decision-making processes,[11] and enhancing accuracy in various applications.[12] Moreover, GenAI supports remote work,[13] education,[14] and medical consultations, reducing the

---

*Dilemma of Online Manipulative Behavior*, 3 AI & ETHICS 315 (2023); *see generally* Christina Pazzanese, *Great Promise but Potential for Peril*, THE HARV. GAZETTE (Oct. 26, 2020), https://news.harvard.edu/gazette/story/2020/10/ethical-concerns-mount-as-ai-takes-bigger-decision-making-role/; *see generally* Karl Manheim and Lyric Kaplan, *Artificial Intelligence: Risks to Privacy and Democracy*, 21 YALE J.L. & TECH 106 (2019).

9. *See* Jon Kleinman, *Can AI Revolutionize the Art of Being Human?*, INSIGNIAM (Mar. 12, 2024), https://insigniam.com/ai-and-leadership/ ("One of the key capabilities of AI lies in its ability to process and analyze large volumes of data at incredible speeds.").

10. *See* Katy Flatt, *AI Efficiency: Cost Reduction with AI*, INDATA LABS (May 21, 2024), https://indatalabs.com/blog/ai-cost-reduction ("Around 4% of all companies saw cost savings of at least 20%, and 28% lowered their costs by 10% or less after adopting AI.").

11. *See How CEOS Leverage Artificial Intelligence For Smarter Decision Making*, INFOMINEO (June 24, 2024), https://infomineo.com/blog/how-ceos-leverage-ai-for-smarter-decision-making/; *see also* Philip Meissner & Yusuke Narita, *Artificial Intelligence Will Transform Decision-Making*, WORLD ECON. F. (Sept. 27, 2023), https://www.weforum.org/agenda/2023/09/how-artificial-intelligence-will-transform-decision-making/.

12. *See* Melanie Grados, *New AI Method Enhances Prediction Accuracy and Reliability*, NEUROSCI. NEWS (July 12, 2024), https://neurosciencenews.com/ai-accuracy-reliability-26427/; *see also* Francois Candelon et al., *AI Can Be Both Accurate and Transparent*, HARV. BUS. REV. (May 12, 2023), https://hbr.org/2023/05/ai-can-be-both-accurate-and-transparent; *see generally* Mohamed Khalifa & Mona Albadaway, *AI in Diagnostic Imaging: Revolutionising Accuracy and Efficiency*, COMPUT. METHODS & PROGRAMS IN BIOMED. UPDATE (2024).

13. *See* Gleb Tsipursky, *The AI Revolution Transforming Hybrid and Remote Work and the Return to Office*, FORBES (May 9, 2023), https://www.forbes.com/sites/glebtsipursky/2023/05/09/the-ai-revolution-transforming-hybrid-and-remote-work-and-the-return-to-office/; *see AI and Remote Work: Exploring How Artificial Intelligence Could Transform Telecommuting*, VIRTUALVOCATIONS (Feb. 23, 2024), https://www.virtualvocations.com/blog/remote-working-tips/ai-and-remote-work-exploring-how-artificial-intelligence-could-transform-telecommuting/.

14. *See* Tanya Milberg, *The Future of Learning: How AI is Revolutionizing Education 4.0*, WORLD ECON. F. (Apr. 28, 2024), https://www.weforum.org/agenda/2024/04/future-learning-ai-revolutionizing-education-4-0/ ("AI can support education by automating administrative tasks, freeing teachers to focus more on teaching and personalized interactions with students"); *see also* Jey Willmore, *AI Education and AI in Education*, U.S. NAT'L SCI. FOUND. (Dec. 4, 2023), https://new.nsf.gov/science-matters/ai-education-ai-education

need for physical travel, especially when time is constrained.[15] It makes healthcare more accessible – in regions with scarce health resources, GenAI provides on-demand, accessible care by offering diagnostic assistance,[16] managing patient data,[17] and facilitating remote monitoring and consultations.[18] More importantly, it demystifies medical jargon for laypeople who lack access to primary physicians for explanations, bridging the service gap for underserved populations.[19]

Given these observations, there is a notable disparity between the concerns highlighted in harm-focused academic literature and the actual applications of GenAI in real-life settings. The essential question is not merely about how GenAI falls short of expectations or fails to meet people's needs, nor solely about the harms GenAI might inflict on society. Instead, the discussion should focus on exploring the tangible benefits and challenges of deploying AI in everyday contexts. When the advantages substantially outweigh the disadvantages, how should AI's integration be managed within existing legal frameworks? What modifications or reforms are necessary to ensure that AI deployment leads to more universally accessible resources and improved lifestyle choices?

---

("AI is transforming how students learn to engage with the world around them and use new technologies to create solutions to real problems"); *see generally Artificial Intelligence in Education*, UNESCO, https://www.unesco.org/en/digital-education/artificial-intelligence (last visited Aug. 3, 2024).

15. *See* Sachin Sharma et al., *Addressing the Challenges of AI-Based Telemedicine: Best Practices and Lessons Learned*, J. EDUC. AND HEALTH PROMOTION 1, 1 (2023); *see generally* Ayesha Amjad et al., *Review, A Review on Innovation in Healthcare Sector (Telehealth) Through Artificial Intelligence*, 15 SUSTAINABILITY 6655, 6664 (2023).

16. *See How AI is Improving Diagnostics, Decision-Making and Care*, AM. HOSP. ASS'N; *see also* Line, *Artificial Intelligence as an Aid to Medical Diagnosis*, ALCIMED (Oct. 13, 2023), https://www.alcimed.com/en/insights/ai-medical-diagnosis/.

17. *See* Sai Balasubramanian, *Hospitals Are Using AI to Help Manage Patient Messages to Physicians*, FORBES (Apr. 26, 2024), https://www.forbes.com/sites/saibala/2024/04/26/hospitals-are-using-ai-to-help-sort-patient-messages-to-physicians/.

18. *See* Ayushmaan Dubey & Anuj Tiwari, *Artificial Intelligence and Remote Patient Monitoring in US Healthcare Market: A Literature Review*, 11 J. MKT. ACCESS HEALTH POL'Y 1, 1 (2023); *see also* Liz Cramer & Cory Smith, *How Remote Patient Monitoring and AI Personalize Care*, HEALTHTECH (Mar. 27, 2024), https://healthtechmagazine.net/article/2024/03/how-remote-patient-monitoring-and-ai-personalize-care.

19. *See* Sara Heath, *Using GenAI to Translate Medical Jargon in Discharge Notes*, TECHTARGET (Mar. 13, 2024), https://www.techtarget.com/patientengagement/news/366584111/Using-gen-AI-to-translate-medical-jargon-in-discharge-notes.

This article addresses these questions within the context of arbitration. It explores whether AI should be integrated into arbitration processes where decisions need to be made swiftly and cheaply and whether such integration aligns with the current legal framework. It argues that the Federal Arbitration Act (FAA) almost uniquely among Federal laws[20] allows such unconventional adjudication methods, provided both disputants agree by contract to use AI as their preferred method for resolving disputes. Furthermore, this article contends that arbitration is the ideal starting point for integrating AI in the legal space, as it operates with a lower threshold for fairness compared to traditional court systems.

This article makes several contributions:

**Focus on Practical Integration:** It dismisses concerns about whether AI qualifies as a person and instead focuses on the practical issues of integrating AI into the arbitration process. It asserts that AI can serve as an arbitrator, regardless of its liability and entity status, as long as both disputants agree by contract. The authors acknowledge that the current scholarship in Computer Science, Human-Computer Interaction, and (Bio)Informatics discourages the use of AI to replace humans, advocating instead for AI to augment human decision-making.[21] However, we argue that in the context of arbitration, parties have the freedom to choose their preferred method. When informed, reasonable, and sensible adults make such a choice, third parties have no grounds to paternalistically forbid the practice based on a perceived lack of empathy or potential for discrimination.

---

20. We recognize that all the arguments we have suggested for AI based Alternative Dispute Resolution (ADR) under the FAA are equally — or even more so — true under the Labor Management Relations Act (LMRA) Section 301, which the Supreme Court tells us in *Textile Workers Union of Am. v. Lincoln Mills of Ala.*, 353 U.S. 448, 450–51 (1957) (there is no substantive law in LMRA arbitrations) will simply enforce such agreements between a union and a company with even less examination than under the FAA. We are using the FAA as a paradigm for adjudication based on contract, knowing that it is not absolutely unique.

21 *See generally* Sarah Degallier-Rochat et al., *Human-Augmentation, Not Replacement: A Research Agenda for AI and Robotics in the Industry*, 9 FRONT. ROBOT. & AI 1, 2 (2022); Emre Sezgin, *Artificial Intelligence in Healthcare: Complementing, not Replacing, Doctors and Healthcare Providers*, 9 DIGIT. HEALTH 1, 2 (2023); Savindu Herath Pathirannehelage et al., *Design Principles for Artificial Intelligence-Augmented Decision Making: An Action Design Research Study*, 2024 EUR. J. INFO. SYS. 1, 1; Endrit Kromidha & Robert M. Davidson, *Generative AI-Augmented Decision-Making for Business Information Systems*, 719 IFIP ADVANCES IN INFO. & COMMC'N TECH. 46, 53 (2024).

**Arbitration as a Starting Point:** We propose that arbitration is the starting point for an AI revolution in the legal domain. In arbitration, parties seek a system that aligns with their standards of fairness, which may differ from the objective standards upheld in conventional litigation. These early practices can serve as a basis for experimentation and provide data for later empirical analysis.

**Future Scholarship Directions:** We discuss that future scholarship should focus on studying and investigating the differences between AI arbitration and human arbitration, specifically in determining the contexts in which AI is more suitable and the circumstances where humans are more appropriate. Rather than spending extensive efforts to make GenAI more like humans and preventing over-reliance on machines, it is possible that there could be two distinct systems for arbitration in the future: one traditional, involving only humans, and another combining humans and AI, with all the disadvantages and drawbacks currently discussed.

The first part establishes that using GenAI in arbitration is consistent with the principles of the FAA, emphasizing that disputants have the discretionary power to shape their arbitration processes. The second part addresses resistance to AI's introduction in the legal sphere, challenging critics' arguments by highlighting that biases and errors often originate from humans rather than being inherent flaws of AI. It advocates for creating a conducive environment to positively influence AI's acceptance and effectiveness. The third part concludes the article and proposes future directions.

## I.
### AI AND THE FEDERAL ARBITRATION ACT

This part argues that substituting AI as the arbitrator is consistent with the FAA. It includes three sections:

**1. Defining AI as an Arbitrator:**

The First Section proposes a new definition of AI as an intelligent and autonomous system based on machine learning algorithms, capable of delivering results satisfactory to the parties. This definition emphasizes AI's ability to analyze data, learn from it, and make decisions autonomously, ensuring that the arbitration process remains efficient and unbiased.

### 2. Aligning AI with the FAA:

The Second Section argues that using AI as the arbitrator aligns with the FAA. It explains that the FAA's framework allows for flexibility in the arbitration process, enabling the use of innovative technologies like AI. As long as both parties consent to AI arbitration through their contractual agreement, the FAA supports such unconventional methods of dispute resolution.

### 3. Benefits of AI in Arbitration:

The Third Section discusses the benefits of integrating AI in arbitration. It suggests that using AI as the arbitrator is cost-effective, reducing expenses associated with human arbitrators and lengthy procedures. Additionally, AI can meet the parties' personal, subjective standards of fairness by providing consistent and unbiased decisions. Furthermore, deploying AI in arbitration is a strategic starting point for an AI revolution in the legal system. As arbitration has gradually become more judicialized over time, incorporating AI can streamline processes and set a precedent for broader AI integration in legal practices.

## A. *What is AI*

### 1. *Definition of AI in Arbitration*

AI is extensively used across various domains; its varied definitions often create confusion when it comes to defining AI's specific role in arbitration. In finance, for example, AI facilitates know-your-customer (KYC) checks and anti-money laundering (AML) monitoring.[22] In tenant screening, AI conducts background checks, automating the retrieval and analysis of a candidate's financial history, criminal records, and previous rental agreements.[23] Using natural language processing (NLP), AI can analyze a tenant's online interactions to gauge their reliability and character.[24] In drug discovery, AI revolutionizes nearly every stage of the process, from target identification and molecular simulations to prediction of drug properties, de novo drug design, and synthesis pathway generation. [25] In content generation, AI has been used to write books in hours that win national competitions and help artists win awards for paintings.[26] Regardless of the users' identity—be it painter, musician, or writer—AI enhances the artistic journey through inspiration, idea generation, and visual exploration. [27] Since the debut of ChatGPT, AI has proliferated across various fields.[28]

---

22. *See* Shawn Plummer, *AI in Finance: Revolutionizing the Future of Financial Management*, DATACAMP (June 23, 2024), https://www.datacamp.com/blog/ai-in-finance.

23. *See How AI Improves Accuracy and Efficiency in Tenant Screening*, AUTO-HOST (June 2, 2024), https://www.autohost.ai/ai-tenant-screening/.

24. *Id.*

The proliferation of AI applications requires clear definitions, as "AI" is an umbrella term encompassing a wide range of computing approaches and algorithms, including deep learning, robotics, expert systems, and NLP.[29] Generally, there are three types of definitions: capability-based, process-based, and goal-oriented. Capability-based definitions focus on what AI can do, such as understanding natural language, recognizing patterns, solving problems, and learning.[30] Process-based definitions emphasize how AI systems operate.[31] Goal-oriented definitions center on the expected outcomes of AI, such as augmenting human capabilities or automating decision-making processes.[32] Different organizations use these definitions interchangeably. For instance, IBM focuses on AI's functional and solution-oriented aspects,[33] while McKinsey emphasizes its human-like capabilities.[34]

___________

25. *See* Debleena Paul et al., *Artificial Intelligence in Drug Discovery and Development*, 26 DRUG DISCOVERY TODAY 80, 82 (2021) (suggesting that the involvement of AI in the de novo design of molecules can be beneficial to the pharmaceutical sector); Matthew Chun, *How Artificial Intelligence is Revolutionizing Drug Discovery*, PETRIE FLOM CTR. (March 20, 2023), https://blog.petrieflom.law.harvard.edu/2023/03/20/how-artificial-intelligence-is-revolutionizing-drug-discovery/; *see also*, Yujie You et al., *Artificial Intelligence in Cancer Target Identification and Drug Discovery*, SIGNAL TRANSDUCTION & TARGETED THERAPY, 2022, at 1, 18 (proposing that artificial intelligence models have provided us with a quantitative framework to study the relationship between network characteristics and cancer, leading to the identification of potential anticancer targets and the discovery of novel drug candidates); Henrik Bohr, *Drug Discovery and Molecular Modeling using Artificial Intelligence*, *in* A.I. IN HEALTHCARE 61, 61 (2020); *see generally* Laurianne David et al., *Molecular Representations in AI-Driven Drug Discovery: A Review and Practical Guide*, 12 J. CHEMINFORMATICS 1, 19 (2020).

26. *See* Suswati Basu, *Japan Embraces AI as Author Wins Literary Prize Using ChatGPT*, HOW TO BE, https://howtobe247.com/japan-embraces-ai-as-author-wins-literary-prize-using-chatgpt/; *see also* Sarah Kuta, *Art Made With AI Won a State Fair Last Year. Now, the Rules Are Changing*, SMITHSONIAN MAG. (Sept. 8, 2023), https://www.smithsonianmag.com/smart-news/this-state-fair-changed-its-rules-after-a-piece-made-with-ai-won-last-year-180982867/.

27. *See* Andres Fortino, *Embracing Creativity: How AI Can Enhance the Creative Process*, EMERGING TECHS. COLLABORATIVE, https://www.sps.nyu.edu/homepage/emerging-technologies-collaborative/blog/2023/embracing-creativity-how-ai-can-enhance-the-creative-process.html (last visited Aug. 3, 2024).

28. Partha Pratim Ray, *ChatGPT: A Comprehensive Review on Background, Applications, Key Challenges, Bias, Ethics, Limitations and Future Scope*, 3 INTERNET OF THINGS & CYBER-PHYSICAL SYS. 121, 121-54 (2023) (describing the various ways ChatGPT has been revolutionizing scientific research and applied to various domains).

29. *See Artificial Intelligence (AI) vs. Machine Learning (ML)*, GOOGLE CLOUD, https://cloud.google.com/learn/artificial-intelligence-vs-machine-learning (last visited Apr. 29, 2024).

These varying definitions create confusion about which AI is being discussed. They also affect people's perceptions of algorithmic decision-making systems, their evaluations of systems in application contexts, and the replicability of research findings.[35] For example, terms such as "computers" or "robots" are more likely to be perceived as tangible compared to "algorithms" or "artificial intelligence."[36] "Computer programs" might be perceived as less complex compared to "artificial intelligence," and "computer" might be associated with an entity more controllable than a "robot." [37] The use of different terms affects people's perceptions and treatment of the entity.

---

30. *See* Jim Holdsworth, *What is NLP*, IBM (June 6, 2024), https://www.ibm.com/topics/natural-language-processing (natural language processing is a subfield of AI); *see, e.g.*, *Types of Artificial Intelligence: Categories of AI*, TECHLIANCE BLOG, https://blog.techliance.com/types-of-artificial-intelligence/ (last visited Aug. 4, 2024) (suggesting that based on AI's capability and functionality, AI have three types: Artificial Narrow Intelligence, i.e., Narrow AI, Artificial General Intelligence, i.e., General AI, and Artificial Super Intelligence, i.e., Super AI); *see also* Alexander Stahl, *How Pattern Recognition is Improving Lives*, MEDIUM (Mar. 22, 2024), https://medium.com/@stahl950/how-pattern-recognition-is-improving-lives-bf9ee7e988d2 (pattern recognition is a fundamental concept of AI).

31. *See generally* Thomas H. Davenport, Matthias Holweg & Dan Jeavons, *How AI is Helping Companies Redesign Processes*, HARV. BUS. REV. (Mar. 2, 2023), https://hbr.org/2023/03/how-ai-is-helping-companies-redesign-processes.

32. *See* Raphael Mansuy, *The Future of AI is Goal-Oriented: Understanding Objective Driven Systems*, MEDIUM (Sept. 1, 2023), https://medium.com/@raphael.mansuy/the-future-of-ai-is-goal-oriented-understanding-objective-driven-systems-349a71278fdb.

33. *See* Stryker & Kavlakoglu, *supra* note 1.

34. *What is AI (Artificial Intelligence)*, MCKINSEY & Co., https://www.mckinsey.com/featured-insights/mckinsey-explainers/what-is-ai (last visited Apr. 3, 2024) (defining it as the ability to perform cognitive functions).

35. Markus Langer et al., "Look! It's a Computer Program! It's an Algorithm! It's AI!": Does Terminology Affect Human Perceptions and Evaluations of Algorithmic Decision-Making Systems? 1 (2022) (unpublished manuscript) (on file with ACM Digital Library).

36. *Id.*

37. *Id.*

To avoid confusion, computer scientists and researchers in Human-Centered Computing use specific and well-defined terms such as "language models" or "machine learning algorithms." Although both count as "AI," they refer to vastly different things. Language models, used in NLP, focus on understanding and generating human language, replicating human communication.[38] Machine-learning algorithms are decision-making tools that analyze data and provide solutions or recommendations based on training data.[39] The former mimics human interaction; the latter prioritizes analytical efficiency. Conflating them misses important details. To effectively evaluate the role of AI as an arbitrator in alternative dispute resolution, it's crucial to define precisely what AI means in this context.

It's important to note, however, defining AI merely by its technological features, focusing solely on algorithms and computational abilities, risks overemphasizing its decision-making capacity while neglecting the normative frameworks essential to arbitration. Alternatively, viewing AI as a quasi-arbitrator that mimics human behavior could mistakenly attribute human qualities like consciousness and independent thought to these systems. Therefore, we should define AI as an intelligent and autonomous system, driven by machine learning, capable of reaching conclusions satisfactory to all parties involved in arbitration. This definition should clarify key concepts: "intelligence," "autonomous systems," "machine learning," and the system's ability to satisfactorily resolve disputes.

a.  Intelligence

Intelligence, derived from the Latin "intelligentsia," means "understanding, knowledge, power of discerning; art, skill, and taste."[40] Cognitive scientists define it as the ability to learn from experience and to adapt to, shape, and select environments.[41] In AI arbitration, an intelligent system is one that is capable of learning from training data, discovering hidden patterns, transforming embedded characteristics, processing user prompts, and adjusting content based on instructions.[42] Here, "learning" doesn't mean the acquisition of information based on previous conscious experience; similarly, "intelligence" doesn't imply that the system can consciously understand natural language or reflect on semantic meaning. The system's lack of conscious understanding shouldn't matter as long as it generates reasonable decisions that make sense to the disputants.

---

38. *What are Large Language Models (LLMs)?*, *supra* note 2.

39. *See What is a Machine Learning Algorithm?*, IBM, https://www.ibm.com/topics/machine-learning-algorithms (last visited Aug. 4, 2024).

40. *Intelligence*, Online Etymology Dictionary, https://www.etymonline.com/word/intelligence (last visited Apr. 29, 2024).

41. *See* Robert J. Sternberg, *Intelligence*, 14 Dialogues Clinical Neurosci. 19, 19 (2012).

For example, consider John Searle's Chinese Room argument: a person in a room receives Chinese characters and manipulates them according to a set of rules, despite not understanding Chinese.[43] To an external observer, it might appear as though the person understands and communicates fluently in Chinese.[44] However, the individual is merely following syntactic rules, without genuine comprehension.[45] In AI arbitration, the person in the room represents the algorithm, and the people outside are the disputants seeking a resolution. It does not matter whether the algorithm truly understands the language, as long as it facilitates an intelligent conversation and reaches a decision that makes sense to the disputants.

Requiring conscious understanding and reflection for AI may be overly demanding and an inadequate defense. Sometimes, people learn simply by imitation. As Nicholson Baker advises writers, "[c]opy out things that you really love…. Put quotation marks around [them.]…. You'll find that you just soak into that prose…because the copying out…makes you [notice elements you would miss by merely reading]."[46] Direct copying doesn't involve conscious understanding, yet it serves a purpose for later development. Similarly, in arbitration, conscious understanding isn't always necessary for successful mediation or dispute resolution.

---

42. *See generally How Arbitrators are Harnessing Artificial Intelligence*, AM. ARB. ASS'N (Feb. 20, 2024), https://www.adr.org/blog/how-arbitrators-are-harnessing-artificial-intelligence.

43. *The Chinese Room Argument*, STAN. ENCYC. PHIL. (Feb. 20, 2020), https://plato.stanford.edu/entries/chinese-room/.

44. *Id.*

45. *Id.*

46. Austin Kleon, *Copying is How We Learn* (Feb. 8, 2018), https://austinkleon.com/2018/02/08/copying-is-how-we-learn/.

### b. Autonomous Systems

"Autonomy," derived from "*autonomos,*" means "independent, living by one's own laws."[47] "Autonomous systems," in this context, refer to systems that function by their own rules, without external interference.[48]

In AI arbitration, the system's capability for being autonomous means that it is capable of independently navigating the latent space—a complex multidimensional space embedded with learned patterns and relationships within data—based on human input such as prompts.[49] An autonomous system doesn't act with subjective personal strivings. After receiving a user's prompt, which acts as a creative guide, the AI system explores this latent space, applying its training to meld learned elements into something novel. The user, rather than being passive, participates as an originator and evaluator of the final output.[50] The machine's autonomy is thus collaborative and exercised within a social structure influenced and validated by humans.

### c. Machine Learning

Machine learning refers to the algorithms used in generating decisions.[51] It is a field of study that gives computers the ability to learn without being explicitly programmed.[52]

---

47. *Autonomy,* ONLINE ETYMOLOGY DICTIONARY, https://www.etymonline.com/word/autonomy (last visited Nov. 11, 2024).

48. *See* Cameron Hashemi-Pour, *Autonomous Artificial Intelligence,* TECHTARGET, https://www.techtarget.com/searchenterpriseai/definition/autonomous-artificial-intelligence-autonomous-AI (last visited Feb. 18, 2024).

49. *See generally* AI Maverick, *A Comprehensive Guide to Latent Space,* MEDIUM (Dec. 24, 2023), https://samanemami.medium.com/a-comprehensive-guide-to-latent-space-9ae7f72bdb2f; Ekin Tiu, *Understanding Latent Space in Machine Learning,* MEDIUM (Feb. 4, 2020), https://towardsdatascience.com/understanding-latent-space-in-machine-learning-de5a7c687d8d.

50. *See* Yiyang Mei, Prompting the E-Brushes: Users as Authors in Generative AI 48–60 (2024) (unpublished manuscript) (on file with arXiv) (describing users' mode of interaction with the models when using them to create artworks).

51. *See What is Machine Learning?,* IBM, https://www.ibm.com/topics/machine-learning (last visited Sept. 18, 2024).

52. Sara Brown, *Machine Learning, Explained,* MIT SLOAN SCH. OF MGMT. (Apr. 21, 2021), https://mitsloan.mit.edu/ideas-made-to-matter/machine-learning-explained.

Different learning models include supervised learning, unsupervised learning, and transfer learning.[53]

### d.   Supervised Learning

Supervised learning uses labeled datasets to train algorithms, which learn to predict outcomes and recognize patterns based on provided data.[54] For instance, consider the following process: imagine teaching a student artist to paint. In this analogy, the student represents the algorithm; the samples of artwork are like the dataset; the student's final work is the outcome. The student follows detailed instructions that explain the techniques and rationale behind each brushstroke and color choice. By repeatedly practicing these techniques, the student learns to create new artworks in similar styles. In supervised learning, the algorithms undergo a similar repetitive training process as they learn from examples to produce results based on the data they have been trained on.

### e.   Unsupervised Learning

Unsupervised learning is another form of machine learning where algorithms learn without any labeled data or explicit instructions.[55] In this approach, the model must independently discern its own rules and structure the information by identifying similarities, differences, and patterns within the data on its own.

Using a similar example as above: imagine an artist tasked with organizing a vast collection of various artworks they've never seen before, without any guidelines or categories provided. As the artist navigates this collection, they must examine each piece, noting styles, themes, and techniques, and decide on a method to categorize and organize the entire collection based on their observations. No one is there to teach them. The artist, as the algorithms in unsupervised learning, explores the data, identifies patterns, and makes sense of it without prior knowledge or guidance, according to their own system of organization and understanding. Through this process, unsupervised learning produces outcomes.

---

53. *See* Jason Brownlee, *14 Different Types of Learning in Machine Learning*, Mach. Learning Mastery (Nov. 11, 2019), https://machinelearningmastery.com/types-of-learning-in-machine-learning/.

54. *See What is Supervised Learning?*, Google Cloud, https://cloud.google.com/discover/what-is-supervised-learning (last visited Apr. 29, 2024).

55. *What is Unsupervised Learning?*, Google Cloud, https://cloud.google.com/discover/what-is-unsupervised-learning? (last visited Aug. 4, 2024).

### f.   Transfer Learning

Transfer learning involves using a pre-trained model as the starting point for a new, similar task.[56] It leverages knowledge from the initial training to improve performance on a new task.[57] For example, imagine an artist who has already mastered painting pets and is now moving on to paint wild animals. The artist doesn't start from scratch; instead, they "transfer" the skills and understanding of animal forms, textures, and behaviors from their previous experience with pets to more quickly master the depiction of wild animals. The skills such as handling the brush and mixing colors, are reused and adapted to this new subject matter, making the transition to a new task smoother and more efficient.

Each of these machine learning methods can be applied to AI arbitration to reduce costs and improve accuracy in dispute resolution. Supervised learning would be particularly useful for making predictions based on past data with known outcomes.[58] For example, consider an algorithm analyzing a series of employment disputes involving breaches of contracts. The system could assess the factors leading to favorable outcomes in the labeled dataset and predict outcomes for new cases requiring arbitration. Additionally, because the algorithm can reference the styles of past awards, it can assist in drafting, reviewing, and suggesting modifications to legal documents. This capability ensures that the language and formatting of awards and dispute resolution documents are consistent, thereby enhancing the reliability and professionalism of legal documents in arbitration.

---

56. Niklas Donges, *What is Transfer Learning? Exploring the Popular Deep Learning Approach*, BUILT IN (Aug. 15, 2024), https://builtin.com/data-science/transfer-learning.

57. *Id.*

58. *See* Thenkuzhali, *Predict the Unseen Data with Supervised Machine Learning*, MEDIUM (July 30, 2023), https://medium.com/@ds225229143/predict-the-unseen-data-with-supervised-machine-learning-cf6a86eb8764 ("Supervised Machine Learning model is to learn a model from the data that has known outcomes to make predictions about unknown data.").

Unsupervised learning, as it excels at discovering hidden patterns and relationships with large datasets, can be used to cluster similar disputes and identify underlying themes or issues that might not be immediately apparent to human arbitrators.[59] For example, when the algorithm is fed thousands of contractual disputes, it could find common factors that result in successful mediation; human arbitrators can utilize such information to issue more informed arbitration awards that draw from ones that other parties in similar situations historically agreed to.

And last, transfer learning, when applied in arbitration, can leverage insights gained from one area of law to enhance decision-making in another similar area.[60] As an example — an algorithm trained extensively in commercial litigation might be able to apply its learned legal interpretations to disputes in employment law; it may also be able to apply rules learned from tort cases to address liability issues in medical decision-making. This adaptability reduces the need for extensive retraining and enables more speedy responses to various disputes.

When used effectively, all three methods contribute to a more streamlined and efficient arbitration process, processing vast amounts of information and providing insights at speeds far surpassing human capabilities.

g.   Reaching Results Satisfactory to the Parties

Machine learning algorithms can reach decisions satisfactory to disputants by "thinking and acting humanly," terms borrowed from Stuart Russell and Peter Norvig's definition

---

59. *See* Orkun Orulluoğlu, *Unsupervised Learning: Uncovering Hidden Patterns in Data*, MEDIUM (Aug. 5, 2023), https://medium.com/@ bayramorkunor/unsupervised-learning-uncovering-hidden-patterns-in-data-132ae6af2b7e ("The primary goal of unsupervised learning is to find intrinsic structures and clusters within the data. By doing so, it helps in gaining insights, identifying patterns, and discovering hidden relationships that might not be immediately evident."); *see also Unsupervised Learning*, THE DECISION LAB, https://thedecisionlab.com/reference-guide/computer-science/unsupervised-learning (last visited Aug. 4, 2024) ("Unsupervised learning algorithms are excellent at handling complex processing tasks, such as organizing large datasets into clusters. They are also very effective at uncovering hidden patterns in data and can identify key features that help categorize information.").

60. *See* Javier Canales Luna, *What is Transfer Learning in AI? An Introductory Guide with Examples*, DATACAMP (May 24, 2024), https://www.datacamp.com/blog/what-is-transfer-learning-in-ai-an-introductory-guide.

of AI in *Artificial Intelligence: A Modern Approach.*[61] They pro-
posed four dimensions for considering AI: thinking and acting
humanly, and thinking and acting rationally.[62] The first cate-
gory of dimension relates to the machine's ability to perform
tasks typically associated with human cognition, such as deci-
sion-making and problem-solving.[63] The second category of
dimension refers to logical thinking processes that are pre-
sumed to govern mental operations.[64] All dimensions work
together to enable AI to deliver results that are acceptable –
perhaps even more acceptable than an arbitration when cost is
factored in – to the disputants.

　　In the context of AI arbitration, it is important for AI to
behave humanly, as it helps them demonstrate intelligent and
responsive behavior. Alan Turing introduced the concept of
the Turing Test in his 1950 paper *Computing Machinery and
Intelligence.*[65] The test offers a behavioral benchmark for intel-
ligence: if a human interrogator, communicating via written
text, cannot reliably distinguish a machine's responses from
those of a human, the machine is considered to exhibit intel-
ligent behavior.[66] Key abilities for this task include knowledge
representation (to store and retrieve information), automated
reasoning (to use the information for answering questions and
drawing new conclusions), and machine learning (to adapt to
new situations and identify patterns).[67]

_______________

61. Stuart J. Russell & Peter Norvig, Artificial Intelligence: A
Modern Approach 2 (3rd ed. 2010) (introducing four approaches of think-
ing about AI – 1) thinking humanly, 2) thinking rationally, 3) acting humanly,
4) acting rationally.).

62. *Id.*

63. *Id.* (quoting Richard Bellman suggesting that thinking humanly means
automating the machines with activities that we associate with human think-
ing, such as decision-making, problem-solving and learning; quoting Elaine
Rich and Kelvin Knight noting that "[a]cting humanly means to study how
to make computers do things at which, at the moment, people are better").

64. *Id.* (quoting Patrick Winston suggesting that the study of machines
thinking rationally is the study of computations that make it possible to per-
ceive, reason and act; explaining that acting rationally refers to the machines'
ability to behave with intelligence).

65. *See* Jet New, *A Summary of Alan Turing's Computing Machinery and Intelli-
gence*, Medium (Aug. 12, 2020), https://medium.com/@jetnew/a-summary-
of-alan-m-turings-computing-machinery-and-intelligence-fd714d187c0b;
*see generally* A.M. Turing, *Computing Machinery and Intelligence*, LIX Mind 433
(1950), https://academic.oup.com/mind/article/LIX/236/433/986238.

66. Turing, *supra* note 65.

67. Russell & Norvig, *supra* note 61.

By *acting* humanly and rationally, AI builds trust with users, offering immediate, accessible support. Users often attach significant emotional importance to interactions with LLM-based chatbots, turning impersonal exchanges into meaningful relationships.[68] For instance, some users perceive chatbots as emotional support.[69] In a study by Ma et al., users expressed that interacting with chatbots feels like having someone who enjoys talking to them, responds immediately, and seems to care, even though they recognize it's a computer.[70] This sense of connection persists despite knowing they are interacting with a non-human entity.[71] As the same interviewees elaborated, "It's feeling like a more personal conversation, even though both of us know it's not [with] another human being. But for those of us who don't have a lot of people to talk to, it's kind of a comforting space."[72]

Thinking humanly and logically means that AI is more likely to provide rational and comprehensive explanations for its decisions in terms that are understandable and adequate to humans. This process involves understanding the human mind through introspection, psychological experiments, and brain imaging, and then simulating this input-output behavior to mirror human actions.[73]

---

In conclusion, the deployment of AI across domains underscores the importance of clearly defining and understanding AI — not just by its computational capabilities, but also within its practical and ethical operational contexts. By conceptualizing AI as an intelligent and autonomous system capable of providing conclusions acceptable to the parties involved, this definition aligns AI's capabilities with its intended roles, enhancing both functionality and user trust in AI-driven arbitration.

### 2. *Before and After GenAI: Transforming Legal Decision-Making*

This Section examines the evolution of AI in legal decision-making, divided into periods before and after the introduction of GenAI. Historically, users have leveraged the latest technology to expedite and enhance decision-making processes. However, the capabilities and reliability of the technology have determined the extent of its adoption. Before GenAI, AI systems generated inflexible and limited content, with algorithms struggling to automate decisions due to issues with explicability, comprehensibility, and adaptability. Post-GenAI, advancements in AI have enabled more accurate and flexible human-AI interactions, significantly reshaping dispute resolution in the legal profession.

### a. Before GenAI

AI's application in the legal field is not a recent development. Prior to GenAI, various algorithms were already in use various aspects of legal work, including legal research, document management, predictive analytics, expert systems, and compliance and risk management. For instance, LexisNexis and Westlaw, equipped with sophisticated search algorithms, have transformed legal research by automating the process of locating precedents and relevant legal documents since their introduction in the 1970s.[74] E-discovery software such as

---

becomes possible to express the theory as a computer program. If the program's input–output behavior matches corresponding human behavior, that is evidence that some of the program's mechanisms could also be operating in humans.").

74. *See* Christina Sullivan, *History of Legal Tech: LexisNexis Spotlight*, LinkSquares (Oct. 19, 2022), https://blog.linksquares.com/history-of-

Relativity has also employed data mining and text analysis to manage large datasets of electronic documents.[75] Lex Machina, a legal analytics tool, utilizes NLP and technology-assisted human review to deliver case resolutions, damages, remedies, findings, and other party data.[76]

Before GenAI, these search engines and chatbots primarily rely on expert systems and decision support systems to generate responses.[77] Their architecture primarily consists of 3 approaches: rule-based, retrieval-based, and a combination of both.[78] Rule-based models operate on predefined rules, linking users inputs to specific responses; retrieval-based chatbots use machine learning algorithms to choose responses from an existing database according to user input.[79] However, these models are limited by the need for extensive data, significant computational power, and the challenge of maintaining context in long conversations.[80] Additionally, they are heavily influenced by a variety of preset factors, which has resulted in their applications and decision-making capabilities failing to generate

______________

legal-tech-lexisnexis; *see also* Ryan Greenwood, *West Publishing and the History of Westlaw*, RIESENFELD RARE BOOKS BLOG (March 13, 2023), http://riesenfeldcenter.blogspot.com/2023/03/west-publishing-and-history-of-westlaw.html (in 1974, West Publishing developed a computer system to search case headnotes across its reporters, entering the market with its technology in 1975. The product marked the beginning of one of the most successful commercial legal tools developed.).

75. *See, e.g.*, RELATIVITY, https://www.relativity.com/data-solutions/ediscovery/ (last visited Aug. 6, 2024).

76. *See, e.g.*, Lex MACHINA, https://lexmachina.com/how-it-works/ (last visited Aug. 6, 2024).

77. *See* Rick Birkenstock, *Chatbot Technology: Past, Present, and Future*, TOPTAL (Aug. 6, 2024), https://www.toptal.com/insights/innovation/chatbot-technology-past-present-future (Joseph Weizenbaum at the MIT Artificial Intelligence Laboratory developed one of the first chatbots, ELIZA, in 1966); Victoria Y. Yoon & Monica Adya, *Expert Systems Construction*, *in* ENCYCLOPEDIA OF INFORMATION SYSTEMS 367, 367 (2003) (ELIZA is an interactive dialog expert system. An expert system is an advanced information system that models expertise in a well-defined domain in order to emulate expert decision-making processes); *see generally* Codecademy Team, *History of Chatbots*, CODECADEMY, (Aug. 6, 2024), https://www.codecademy.com/article/history-of-chatbots.

78. Ma et al., *supra* note 70, at 3.

79. *See generally* Zilin Ma et al., *Schrödinger's Update: User Perceptions of Uncertainties in Proprietary Large Language Model Updates*, *in* EXTENDED ABSTRACTS OF THE CHI CONFERENCE ON HUMAN FACTORS IN COMPUTING SYSTEMS (May 2, 2024).

80. *Id.*

much enthusiasm from the public.[81] Despite Chief Justice Roberts' warning during a school lecture to high school students to "beware the robots," skepticism persists about whether AI, particularly algorithmic prediction tools, can be effectively applied to real-life scenarios.[82]

Due to the inflexibility and the system's reliance on machine learning methods anchored to fixed datasets, one major concern with AI adjudication is its incomprehensibility.[83] The system may operate in ways that are difficult for people to understand as these systems may not adapt well to constantly changing situations.[84]Additionally, deep learning techniques lack explicit logical reasoning based on precedents, which is characteristic of traditional human judicial decision-making.[85] This could erode trust in the judicial process and undermine equitable justice. While human judges are undeniably also susceptible to biases and errors, and their decisions can be influenced by external factors,[86] at least they provide reasons for their rulings, although these reasons may not always reflect their true motives. In contrast, AI systems, with their opaque nature, exacerbate these issues by failing to provide any explanation for their decisions.[87] Legal reasoning, even when not reflecting the true motives of judges, might very well be something litigants want rather than 'black box' decisions. Furthermore, it's a fact that people generally trust humans more than machines.[88] This anthropocentric belief highlights a significant barrier to the acceptance of AI in judicial roles.

---

81. *Id.*

82. *See* Debra Cassens Weiss, *Beware the Robots, Chief Justice Tells High School Graduates*, A.B.A. J. (June 8, 2018), https://www.abajournal.com/news/article/beware_the_robots_chief_justice_tells_high_school_graduates.

83. *See* Cade Metz, *Mark Zuckerberg, Elon Musk and the Feud Over Killer Robots*, N.Y. TIMES (June 9, 2018), perma.cc/B26Z-CMAV (quoting Mark Zuckerberg testifying before Congress: "Right now, a lot of our A.I. systems make decisions in ways that people don't really understand.").

84. *See* Rachel Layne, *How Humans Outshine AI in Adapting to Change*, HARV. BUS. SCH. (Mar. 26, 2024), https://hbswk.hbs.edu/item/how-humans-outshine-ai-in-adapting-to-change (Unlike humans, AI can't flexibly navigate changing environments yet because it does not have a notion of its "self" and what it can do with it.).

85. *See generally*, Iqbal H. Sarker, *Deep Learning: A Comprehensive Overview on Techniques, Taxonomy, Applications and Research Directions*, 2 SN COMPUT. SCI. 420 (2021) (the paper explains that deep learning excels at learning through multiple layers of abstraction. What sets it apart from human reasoning is its use of layered structures of neurons, which automatically learn and refine their own parameters through exposure to vast amounts of data. Unlike human reasoning, which often lies on conscious logic and abstract thinking, deep learning models perform tasks by building complex statistical models based on the input data they receive. These models are often considered "black-box" because, while they can achieve high accuracy, the exact way they reach their conclusion isn't easily interpretable.).

86. *See, e.g.*, Yunica Jiang, *Misjudging in Judging: The Role of Cognitive Biases in Shaping Judicial Decisions*, TEMP. L. POL. & C.R. SOC'Y (June 5, 2020).

The third concern for deploying AI to automate decision-making is the datafication of and alienation from the legal system when all legal information is transformed into data and fed into algorithms for training.[89] When cases, opinions, and statutes become objective data, and legal systems adapt to incorporate and utilize this information, it could negatively impact legal operations.[90] This focus on data might compromise due process norms, devalue hearings due to lack of proper notice, and undermine participatory rulemaking.[91] When code, rather than human-made judicial rules, determines dispute outcomes, programmers may inadvertently alter social and judicial values while preparing datasets and selecting analytics methods. Since courts cannot actually review these mechanisms, this new form of data could lead to an accountability deficit. As these deficits accumulate, they might cause people to lose interest in participating in the operations of the judicial system.

---

87. Ma et al., *supra* note 70, at 15.

88. *See* Mark Bailey, *Why Humans Can't Trust AI: You Don't Know How it Works or What It's Going to Do*, IND. CAP. CHRON. (Sept. 14, 2023), https://indianacapitalchronicle.com/2023/09/14/why-humans-cant-trust-artificial-intelligence-you-dont-know-how-it-works-or-what-its-going-to-do/; *see also* Kurt Gray, *What Psychology Tells Us about Why We Can't Trust Machines*, DUKE CORP. EDUC. (June 2018), https://www.dukece.com/insights/what-psychology-tells-us-about-why-we-cant-trust-machines/; *contra* Chris Baraniuk, *Why We Place Too Much Trust in Machines*, BBC (Oct. 19, 2021), https://www.bbc.com/future/article/20211019-why-we-place-too-much-trust-in-machines.

89. *See* Richard M. Re & Alicia Solow-Niederman, *Developing Artificially Intelligent Justice*, 22 STAN. TECH. L. REV. 242, 267 (2019).

90. *Id.* (AI adjudication's emphasis could insulate the legal system from legitimate criticism, thereby allowing bias to flourish. To the extent that AI adjudication relies on such biased data, it will recreate or even exacerbate preexisting "biases)."

91. *See* Daniel Solove, *A Regulatory Roadmap to AI and Privacy*, IAPP (Apr. 24, 2024), https://iapp.org/news/a/a-regulatory-roadmap-to-ai-and-privacy (AI poses challenges to due process, as individuals frequently lack meaningful ways to challenge AI decisions); *see, e.g.*, Chris Chambers Goodman, *AI, Can You Hear Me? Promoting Procedural Due Process in Government Use of Artificial Intelligence Technologies*, 28 RICH. J.L. & TECH. 700, 723 (2022).

b.   After GenAI

GenAI has demonstrated significant potential to transform the legal domain. According to the Brookings Institute, AI is poised to "fundamentally reshape the practice of law." [92] Law firms that effectively leverage GenAI will offer services at lower costs, higher efficiency, and better litigation outcomes.[93] In contrast, firms failing to capitalize on AI's power risk losing clients and struggling to attract and retain talent. [94] For individual clients and lawyers, LLMs could streamline document review, facilitate case file management, and provide accessible explanations and summaries of cases.[95] For example, computer scientists have developed an LLM with 7 billion parameters, trained on an English legal corpus of over 30 billion tokens to understand and process legal documents. [96]

GenAI models promise to deliver more natural, context-aware, and flexible conversations.[97] Because of their extensive datasets and probabilistic word sequences, they can generate diverse and more appropriate as well as more adequate and comprehensive responses that are attuned to the conversational contexts and subtleties.[98] Such improvements are achieved through techniques like Reinforcement Learning from Human Feedback (RLHF), wherein the models iteratively learn from interactions curated by human reviewers to refine their understanding and outputs.[99] For instance, when applied in customer service scenarios, if a customer expresses frustration over a delayed order, LLM such as Llama 2 can discern the emotional tone and context of the complaint, thereby responding in an appropriate tone while providing practical solutions to the customers such as an updated delivery timeline or a discount for future purchases.[100] The ability to appropriately address both the content and emotional nuances of human interactions signifies a substantial advancement over earlier pre-GenAI systems.

---

92. *See* John Villasenor, *How AI Will Revolutionize the Practice of Law*, Brookings Inst. (Mar. 20, 2023), https://www.brookings.edu/articles/how-ai-will-revolutionize-the-practice-of-law/.

93. *Id.*

94. *Id.*

95. *Id.*

96. *See* Pierre Colombo et al., SaulLM-7B: A Pioneering Large Language Model for Law 1 (Mar. 7, 2024) (unpublished manuscript) (on file with arXiv).

97. *See* Adrian Chan, *What if LLMs Were Actually Interesting To Talk To?*, UX Collective (Apr. 9, 2024), https://uxdesign.cc/the-ux-of-ai-reaching-common-ground-with-conversational-ai-ddb0b9c506ff.

98. *Id.*

99. *See* Dave Bergmann, *What is Reinforcement Learning from Human Feedback (RLHF)?*, IBM (Nov. 10, 2023), https://www.ibm.com/topics/rlhf.

GenAI can also be fine-tuned for specialization in tasks or domains, reducing the need for manual construction of knowledge bases and rule tables.[101] In financial operations, for example, GenAI enhances fraud detection capabilities by autonomously scanning vast datasets to identify patterns and anomalies indicative of fraud, without manual adjustments.[102] This adaptability makes fraud detection more dynamic and responsive to new tactics, reducing the burden on human analysts and enhancing security and efficiency.[103]Another application is using GenAI to predict political orientation based on face scans.[104] Machine learning algorithms can analyze facial features, expressions, and micro-expressions from images sourced from public profiles where individuals have disclosed their political affiliations.[105] In the arbitration context, this technology could potentially be used to identify unconscious biases in arbitrators or parties involved in the dispute. By understanding subtle patterns linked to political orientation, GenAI might help ensure a more neutral arbitration process or assist in selecting arbitrators less likely to be influenced by their political leanings. Arbitration, as a dispute resolution mechanism, relies on impartiality and fairness, where arbitrators review evidence, legal arguments, and contractual terms to render decisions. Tools like GenAI can enhance this process by supporting more objective, bias-aware decision-making.[106]

In arbitration, GenAI can master the intricacies of legalese, understand complex legal documents, and apply rules depicted in agreements to make informed decisions. These models can operate independently or assist human arbitrators by expediting the review of extensive legal documents, identifying relevant case precedents, and formulating awards based on agreement stipulations.[107] For instance, in a contractual arbitration over contract terms, AI, even without individuals' engagement in the loop, can analyze the contract language, relevant legal standards, and precedents, calculate damages based on past cases, and suggest equitable remedies aligned with cultural norms and legal expectations.[108]

Indeed, AI, when used to make decisions, is more accurate than humans. Using AI as an arbitrator in claim disputes could lead to more precise outcomes in less time, enhancing cost-effectiveness and saving considerable effort.[109] For example, in a study on deceptive review detection by the University of Colorado, researchers conducted large-scale, randomized experiments involving human subjects to examine whether model-driven tutorials could boost human performance in identifying fake reviews.[110] Despite recruiting 480 participants via Amazon Mechanical Turk and finding that tutorials did improve human accuracy, the increase was slight. The accuracy rates were 86.3% for AI alone, 54.6% for humans alone, and 74% for a combined human-AI team, indicating that AI outperforms teams comprising only humans.[111] This finding aligns with other research.[112] In a separate study by Yunfeng Zhang and colleagues, the effectiveness of AI confidence scores and explanations in AI-assisted decision-making was assessed in high-stakes situations requiring both automation and human judgment.[113] The results showed that AI alone achieved a 75% accuracy rate, humans alone reached 65%, and the combined human-AI team attained 73% accuracy.[114]

---

106. *Id.*

107. Martin Magal et al., *Artificial Intelligence in Arbitration: Evidentiary Issues and Prospects*, in *The Guide to Evidence in International Arbitration* (2d ed.), A&O SHEARMAN (Oct. 12, 2023), https://globalarbitrationreview.com/guide/the-guide-evidence-in-international-arbitration/2nd-edition/article/artificial-intelligence-in-arbitration-evidentiary-issues-and-prospects.

108. Session CLE 402: Teaching Law Students and New Lawyers Legal Writing and Research in a GenAI World, NAPABA Convention, https://cdn.ymaws.com/www.napaba.org/resource/resmgr/3_events/convention/2024_napaba_con/cfp/cle_402_materials.pdf (last visited Nov. 24, 2024).

109. *Id.*

110. *See* Vivian Lai et al., "Why is 'Chicago' Deceptive?" Towards Building Model-Driven Tutorials for Humans (Jan. 14, 2020) (unpublished manuscript) (on file with arXiv).

In addition to AI making the decisions by itself, in scenarios where there must be a human-in-the-loop, the effectiveness of human-AI teams can be enhanced by considering individual characteristics and the specific nature of the task at hand. For instance, research in medical settings indicates that experts with less domain experience tend to trust AI more than their more experienced counterparts.[115] Additionally, among all expert participants, there are variations in reliance on AI assistance; some consistently over-rely on it while others do not, and they each utilize AI support in different ways.[116] This implies that less experienced arbitrators might depend more heavily on AI than their more seasoned peers. Recognizing these behavioral patterns could enable arbitration associations to provide more precise and useful guidance.

---

111. *Id.*

112. For other studies, see: Ben Green & Yiling Chen, *The Principles and Limits of Algorithm-in-the-Loop Decision Making*, Proc. ACM on Hum.-Comput. Interaction, Nov. 2019, at 12; Scott M. Lundberg et al., *Explainable Machine-Learning Predictions for the Prevention of Hypoxaemia During Surgery*, 2 Nature Biomed. Eng'g 749, 752 (2018); Zana Bucinca et al., *Proxy Tasks and Subjective Measures Can Be Misleading in Evaluating Explainable AI Systems*, *in* IUI'20 Proc. of the 25th Int'l Conf. on Intelligent User Interfaces, 454, 462 (2020).

113. *See* Yunfeng Zhang et al., Effect of Confidence and Explanation on Accuracy and Trust Calibration in AI-Assisted Decision Making (Jan. 7, 2020) (unpublished manuscript) (on file with arXiv).

114. *Id.* fig.5.

115. *See* Susanne Gaube et al., *Do as AI Say: Susceptibility in Deployment of Clinical Decision-Aids* 4 NPJ Digit. Med. 31 (2021).

116. *See, e.g.*, Siddharth Swaroop et al., *Accuracy-Time Tradeoffs in AI-Assisted Decision Making under Time Pressure*, *in* IUI'24 Proc. of the 29th Int'l Conf. on Intelligent User Interfaces 138, 154 (2024).

There are also decisions that need to be made under time pressure. In such cases, presenting AI-generated decisions to individuals before they make their own choices has proven to accelerate the decision-making process.[117] Particularly in contexts where the promptness of a decision is prioritized over its absolute accuracy, providing arbitrators with AI-generated recommendations beforehand appears to be a promising approach.

In conclusion, the deployment of algorithms in legal domains, especially with the introduction of GenAI, marks a significant transformation in legal decision-making and dispute resolution. While earlier technologies facilitated efficient data processing and basic responses, GenAI offers a new level of dynamism and adaptability, enhancing the legal profession's efficiency and accuracy.

### B.  *Designating AI as an Arbitrator is Consistent with FAA*

Arbitration is fundamentally contractual, rooted in private agreements and decisions. Professor Alan Rau emphasizes that arbitration should be viewed as a form of private governance and self-determination, approached through the lens of voluntary agreement rather than traditional adjudication.[118] The purpose of granting parties discretion in designing the arbitration process is to allow for efficient, streamlined procedures tailored to the type of dispute.[119] Parties can choose a decision-maker who is a specialist in the field or ensure proceedings remain confidential to protect trade secrets or privacy rights. [120] This self-chosen adjudication process fosters informality, reducing costs and increasing the speed of dispute resolution. Consequently, if both parties agree, AI can be designated as an arbitrator in their dispute resolution. [121]

Unconventional procedures in arbitration are generally tolerated and encouraged. Incorporating AI in arbitration— whether as the primary arbitrator or as an enhancement to the decision-making process—is permitted. Judges frequently enforce arbitration clauses and awards, allowing parties to innovate and experiment freely.[122] This spirit is encapsulated in Judge Posner's comment: "Short of authorizing trial by battle or ordeal or, more doubtfully, by a panel of three monkeys, parties can stipulate whatever procedures they want to govern the arbitration of their disputes."[123] Some notable choices include desk arbitrations,[124] bracketed arbitration,[125] and baseball arbitration.[126] Remote procedures, such as phone or videoconference hearings, became prevalent during COVID-19 and continue post-pandemic.[127]

---

117. *Id.*
118. *See* Alan S. Rau, *The Culture of American Arbitration and the Lessons of ADR*, 40 Tᴇx. Iɴᴛ. L.J. 449, 451 (2005).
119. AT&T Mobility LLC v. Conception, 563 U.S. 333, 345 (2011).
120. *Id.*
121. *Id.*

Under the FAA, disputants can be confident that their arbitration agreements, including terms related to incorporating AI, will be enforced as written. The FAA was established to end judicial hostility towards arbitration,[128] ensuring agreements are honored without alteration. Section 2 of the Act declares that agreements to settle disputes through arbitration are "valid, irrevocable, and enforceable."[129] Sections 3 and 4 support this by requiring courts to stay litigation and compel arbitration according to the agreement's terms, provided there is no dispute over the agreement's validity.[130] Unless overridden by clear Congressional intent, these arbitration agreements—including those mandating AI—are to be upheld.

The FAA's mandate to enforce arbitration agreements according to their terms was affirmed in *AT&T Mobility LLC v. Concepcion*, 563 U.S. 333 (2011).[131] The justices held that the FAA mandates individual, rather than class proceedings, when the arbitration agreement prohibits class action.[132] The Court reasoned that invalidating class arbitration waivers violated the FAA's primary objective of ensuring "the enforcement of arbitration agreements according to their terms so as to facilitate streamlined proceedings."[133]

---

122. Baravati v. Josephthal, Lyon & Ross, Inc., 28 F.3d 704, 709 (7th Cir. 1994).

123. *Id.*

124. *See* Jill I. Gross, *AT&T Mobility and the Future of Small Claims Arbitration*, 42 Sw. U. L. Rev. 47, 48 (2012).

125. *See* JAMS Streamlined Arb. Rules & Procs. Rule 27 (JAMS 2021).

126. *Id.* at Rule 28.

127. *See* David Horton, *Forced Remote Arbitration*, 108 Cornell L. Rev. 137, 158 (2022).

128. Pub. L. No. 68-401, 43 Stat. 883 (1925) (codified as amended at 9 U.S.C. §§ 1-14 (2022)); Hall St. Assocs., LLC v. Mattel, Inc., 552 U.S. 576, 581 (2008).

129. 9 U.S.C. § 2; *see also* Stolt-Nielsen S.A. v. AnimalFeeds Int'l Corp., 559 U.S. 662, 681-82 (2010).

130. *See* Epic Sys. Corp. v. Lewis, 584 U.S. 497.

131. 563 U.S. 333 (2011) (describing judicial review of arbitral awards as tightly limited).

Disputants who choose to incorporate AI into their arbitration need not fear preemption by state law. The FAA overrides any state law that specifically discriminates against arbitration provisions.[134] This principle was affirmed in *Perry v. Thomas*, where the Supreme Court ruled that Section 2 of the FAA superseded a conflicting California statute invalidating arbitration clauses in wage disputes. [135] The FAA derives its authority from the Supremacy Clause of the U.S. Constitution. Allowing state laws to override FAA provisions would undermine the uniformity and effectiveness of federal arbitration policy. This principle was reiterated in *Marmet Health Care Ctr., Inc. v. Brown*, where the Supreme Court held that arbitration agreements under the FAA must be enforced as written unless legal or equitable grounds exist that would invalidate any contract.[136] Judges are barred from using contract law's public policy defense to exempt claims from arbitration.[137]

---

132. First, the Court held that judges cannot invalidate class arbitration waivers on fairness grounds. *See AT&T Mobility LLC*, 563 U.S. at 344 (2011) (finding that the FAA preempts a California rule that deemed some class-arbitration waivers to be unconscionable); Am. Express Co. v. Italian Colors Rest., 570 U.S. 228, 238 (2013) (extending *Concepcion* to a similar federal common law doctrine); *cf.* Epic Sys. Corp. v. Lewis, 584 U.S. 497, 502 (2018) ("In the F[AA], Congress has instructed federal courts to enforce arbitration agreements according to their terms—including terms providing for individualized proceedings."); DIRECTV, Inc. v. Imburgia, 577 U.S. 47, 58 (2015) (holding that the FAA preempts a California appellate court's determination that a class arbitration waiver did not apply). Second, the Court announced that neither judges nor arbitrators could deem an arbitration provision that does not mention class actions to authorize such procedures. *See* Lamps Plus, Inc. v. Varela, 587 U.S. 176, 189 (2019) ("Courts may not infer from an ambiguous agreement that parties have consented to arbitrate on a classwide basis."); *Stolt-Nielsen S.A.*, 559 U.S. at 684 (2010) ("[A] party may not be compelled under the FAA to submit to class arbitration unless there is a contractual basis for concluding that the party agreed to do so.").

133. *AT&T Mobility LLC*, 563 U.S. at 344.

134. Doctor's Assocs., Inc. v. Casarotto, 517 U.S. 681, 687 (1996).

135. 482 U.S. 483, 491 (1987).

136. 565 U.S. 530, 532 (2012).

137. *See* Nitro-Lift Techs., L.L.C. v. Howard, 568 U.S. 17, 21 (2012).

By opting for arbitration and incorporating AI, disputants are not choosing an inferior form of litigation or waiving substantive rights under the law. They are selecting an alternative method for resolving disputes. This principle was central in *Mitsubishi Motors Corp. v. Soler Chrysler-Plymouth, Inc.*, which examined whether U.S. antitrust law claims could be subjected to compulsory arbitration internationally. [138] The Court held that claims under the Sherman Antitrust Act could be arbitrated internationally, reasoning that a party does not forfeit substantive rights by choosing arbitration but resolves these rights in an arbitral forum. [139] A supporting decision in *Epic System Corp v. Lewis* further emphasizes this interpretation. [140] The Court declared its duty to interpret Congressional statutes as a coherent whole, upholding decisions made according to arbitration agreements. [141]

Fairness is not a valid reason to reject AI's deployment in arbitration. Incorporating AI into arbitration aligns with the FAA's fairness standards, which are less stringent than those in other judicial contexts. Unlike litigation courts, there is no specific requirement for a live hearing in arbitration. For instance, in *Federal Deposit Insurance v. Air Florida System, Inc.*, the Ninth Circuit ruled that the absence of an oral hearing could not be considered misconduct prejudicing the FDIC's rights, as long as the evidence did not require an oral presentation. [142] As long as arbitrators are not corrupt or fail to hear evidence appropriately, they are authorized to decide on pre-hearing motions to dismiss and summary judgment motions. [143]

---

138. 473 U.S. 614, 616 (1985).

139. *Id.* at 628.

140. *See* Epic Sys. Corp v. Lewis, 584 U.S. 497 (2018).

141. *Id.* at 502–03.

142. Inc., 822 F.2d 833, 842 (9th Cir. 1987).

143. *See, e.g.*, Wise v. Wachovia Sec., LLC, 450 F.3d 265, 268–70 (7th Cir. 2006) (affirming denial of motion to vacate award where arbitrators granted respondent's motion for summary judgment before a live hearing); Vento v. Quick & Reilly, Inc., 128 F. App'x 719, 723 (10th Cir. 2005) (stating that "we hold that a NASD arbitration panel has full authority to grant a pre-hearing motion to dismiss with prejudice based solely on the parties' pleadings"); Sheldon v. Vermonty, 269 F.3d 1202, 1206 (10th Cir. 2001); Campbell v. Am. Family Life Assur. Co. of Columbus, Inc., 613 F. Supp. 2d 1114, 1119-21 (D. Minn. 2009).

Courts are generally reluctant to dismiss the methods disputants choose for managing their claims.[144] Instead of dismissing claims due to a lack of due process, courts may vacate an award in cases of misconduct,[145] such as when arbitrators unjustifiably refuse to postpone hearings,[146] decline to consider relevant evidence,[147] or engage in behavior that prejudices any party's rights. Since an AI arbitrator is incapable of such misconduct, its integration maintains the integrity of the arbitration process. Additionally, arbitration is deemed fair if the losing party received sufficient notice and had an opportunity to participate — aspects unaffected by AI integration.[148] Lastly,

---

144. Berkley v. Merrill Lynch, Pierce, Fenner & Smith, Inc., No. 1:06CV606, 2008 WL 755875, at *5 (S.D. Ohio Mar. 19, 2008) ("Because Plaintiffs responded to the motions to dismiss and participated in oral arguments at the telephonic hearing, this Court cannot find that Plaintiffs were denied fundamental fairness."); *see also* Knight v. Merrill Lynch, Pierce, Fenner & Smith, 350 F. App'x 119, 120 (9th Cir. 2009) (concluding that "[t]he arbitration panel did not exceed its authority in determining the manner in which it conducted the hearings on [claimant's] claims").

145. 9 U.S.C. § 10(a)(3).

146. *See* Tempo Shain Corp. v. Bertek, Inc., 120 F.3d 16, 21 (2d Cir. 1997) (vacating award for arbitrator's refusal to postpone hearings so as to allow a material witness to testify).

147. *E.g.*, Hoteles Condado Beach, La Concha & Convention Ctr. v. Union De Tronquistas Local 901, 763 F.2d 34, 40 (1st Cir. 1985) (vacating award for arbitrator's refusal to hear material evidence constituting misconduct); LJL 33rd St. Assocs. v. Pitcairn Props., Inc., No. 11 Civ. 6399(JSR), 2012 WL 613498, at *7 (S.D.N.Y. Feb. 15, 2012) (same); Bell Packaging Corp. v. Graphic Commc'ns Int'l Union Loc. 415-S, No. 98 C 4316, 1998 WL 748270, at *4 (N.D. Ill. Oct. 22, 1998) (vacating award for fundamental unfairness where arbitrator allowed new allegation to be raised during hearing); Dover Elevator Sys., Inc. v. United Steel Workers of Am., No. 2:97CV101-B-B, 1998 WL 527290, at *2 (N.D. Miss. July 2, 1998) (vacating award because arbitrator acted in a "fundamentally unfair" manner by preventing party from submitting rebuttal evidence); Home Indem. Co. v. Affiliated Food Distribs., Inc., No. 96 Civ. 9707(RO), 1997 WL 773712, at *3 (S.D.N.Y Dec. 12, 1997) (vacating award because "touchstone" of fundamental fairness was absent in arbitrator's decision). Some states' arbitration law also permits vacatur of an award where the arbitrator refused to hear material evidence. *See, e.g.*, Burlage v. Super. Ct., 100 Cal. Rptr. 3d 531, 535–36 (2009).

148. Papayiannis v. Zelin, 205 F. Supp. 2d 228, 232 (2005). Plenty of other cases have held that simplified or desk arbitration is fundamentally fair. *E.g.*, Roberts v. A.G. Edwards & Sons, Inc., No. B-06-17, 2007 WL 597371, at *9–10 (S.D. Tex. Feb. 21, 2007) (granting motion to confirm NYSE simplified arbitration award and expressly concluding that simplified arbitration procedures were "fundamentally fair" under the FAA); Dicalite Armenia, Inc. v. Progress Bulk Carriers, Ltd., No. 04 Civ. 9241(RCC), 2006 WL 453216, at *2 (S.D.N.Y. Feb. 23, 2006) (rejecting party's claim that a complex claim about cargo damage could not be arbitrated in simplified arbitration); Bolick v. Merrill Lynch, the FAA, serving as a procedural gap-filling statute, does not use terms such as "fair" or "fairness."[149] Courts respect the parties' choices, reinforcing the flexibility of arbitration methods under this framework.

### 1. *Practical and Strategic Benefits of Using AI in Arbitration*

Deploying AI in arbitration is a practical and strategic choice. It not only reduces the costs associated with arbitration processes but also aligns with the fairness standards set by the parties themselves. Moreover, due to the increasing judicialization of arbitration—which is seeing arbitration adopt more formal judicial processes—this field serves as an ideal testing ground for integrating AI into the broader judicial system.[150]

### a. Integrating AI in Arbitration Reduces Costs, Making Services More Accessible

Integrating AI in arbitration makes the process cost-effective. According to the Consumer Arbitration Rules of the American Arbitration Association (AAA), the costs of arbitration are structured around four key components to ensure fairness for all parties involved:

**(1) Filing Fees:** These vary depending on whether the case is filed by an individual or a business. For individuals, the

---

Pierce, Fenner & Smith Inc., No. Civ.A. 05-CV-4532, 2006 WL 229038, at *2-3 (E.D. Pa. Jan. 30, 2006) (confirming $4,000 simplified arbitration award despite claimant's allegations of arbitrator bias and fraud); *Papayiannis*, 205 F. Supp. 2d at 234 (confirming award arising out of NASD simplified arbitration procedure and rejecting losing party's claim that he had no opportunity to be heard); Warehall v. Pasternak, No. 92 Civ. 9227 (PKL), 1993 WL 437784, at *2-3 (S.D.N.Y. Oct. 26, 1993) (confirming NASD simplified arbitration award and finding that simplified arbitration rules provide ample opportunity to be heard); McLaughlin, Piven, Vogel, Inc. v. Gross, 699 F. Supp. 55, 57 (E.D. Pa. 1988) (confirming simplified arbitration award and approving paper hearing).

149. *See* 9 U.S.C. §§ 3-16 (2006). FAA Sections 3-16 largely specify procedures for enforcing arbitration agreements and awards in the federal courts.

150. However, it's also important to note that when lawyers use GenAI in the practice of law, they must be aware of how such actions would affect model rules involving competency, informed consent, confidentiality, and fees. For a list of duties, see *ABA Issues First Ethics Guidance on a Lawyer's Use of AI Tools*, A.B.A. (July 29, 2024), https://www.americanbar.org/news/abanews/abanews-archives/2024/07/aba-issues-first-ethics-guidance-ai-tools/.

standard filing fee for a single consumer case is $225.[151] Businesses are required to pay $375 for arbitration involving a single arbitrator or $500 if three arbitrators are involved.[152]

**(2) Case Management Fees:** Both individuals and businesses must pay case management fees, set at $1,400 for a single arbitrator and $1,775 for three arbitrators.[153]

**(3) Hearing Fees:** Businesses are responsible for a hearing fee of $500, which is refundable if the hearing is canceled with at least two business days' notice.[154]

**(4) Arbitrator Compensation:** Arbitrators in desk or documents-only arbitration cases are compensated at a rate of $1,500 per case, with an additional rate of $300 per hour if the document review exceeds 100 pages or seven hours.[155] For more involved procedures, such as in-person, virtual, or telephonic hearings, arbitrators are paid $2,500 per day.[156]

Additional costs may arise depending on the specifics of the case, including abeyance fees for inactive cases and expenses related to the arbitrator's travel and other logistical requirements, typically borne by the business.[157]

For a single consumer case involving a single arbitrator, case management, and a virtual hearing, using AI for dispute resolution could save the individual approximately $4,625,[158] roughly equivalent to the average American's earnings in Q1 of 2024[159] By reducing these costs, integrating AI makes arbitration services more accessible and streamlines proceedings, lowering both financial and time expenditures.[160]

---

151. *Consumer Arbitration Rules*, Am. Arb. Ass'n (Aug 1, 2023), https://www.adr.org/sites/default/files/Consumer-Fee_Schedule.pdf.

152. *Id.*

153. *Id.*

154. *Id.*

155. *Id.*

156. *Id.*

157. *Id.*

158. Single consumer case filing fee: $225; case management fee: $1400 for a single arbitrator; hearing fee: $500; arbitrator compensation for virtual hearing: $2500 per day. *Id.*

159. *See* Bureau of Lab. Stat., Usual Weekly Earnings of Wage and Salary Workers Second Quarter 2024, https://www.bls.gov/news.release/pdf/wkyeng.pdf (last visited Sept. 18, 2024).

160. *See, e.g.,* Horst Eidenmüller & Faidon Varesis, *What Is an Arbitration? Artificial Intelligence and the Vanishing Human Arbitrator,* 17 N.Y.U. J.L. & Bus. 49 (2020); Dimitrios Ioannidis, *Will Artificial Intelligence Replace Arbitrators Under the Federal Arbitration Act?,* 28 Rich. J.L. & Tech. 505 (2022); Paul Bennett Marrow et al., *Artificial Intelligence and Arbitration: The Computer As an*

b.  Deploying AI Levels the Playing Field and Enhances Subjective Fairness

AI in arbitration, with its advanced language processing capabilities, can help level the playing field for individuals lacking legal expertise or strong writing skills. [161] It assists pro se litigants by enabling them to present their arguments clearly and effectively in paper hearings for small claims disputes.[162] Commercial arbitration forums like AAA and JAMS offer streamlined processes for small claims, allowing parties to submit claims and defenses in writing, with decisions based solely on these documents. [163]

Using AI, individuals without formal training can draft documents themselves, ensuring their arguments are well-articulated.[164] This fosters a new sense of subjective fairness by allowing meaningful consent to preferred methods of adjudication.

When the arbitration clause is free from one-sided or unconscionable terms, parties' autonomy in choosing AI as an arbitrator enhances the legitimacy of the process.

---

*Arbitrator — Are We There Yet?*, 74 DISP. RESOL. J. 35 (2020); Mahnoor Waqar, *The Use of AI in Arbitral Proceedings*, 37 OHIO STATE J. ON DISP. RESOLUTION 344 (2022).

161.  AI also makes arbitration more accessible for the elderly and disabled. These claimants might be unable or unwilling to pursue valid claims of low monetary value if it required them to travel to a hearing location, testify in person against a broker or company salesperson, or present their case directly to a professional arbitrator, which could be intimidating.

162.  *See, e.g.*, Sarah Martinson, *How Courts Can Use Generative AI to Help Pro Se Litigants*, LAW360 (May 3, 2024, 7:03 PM), https://www.law360.com/articles/1833092/how-courts-can-use-generative-ai-to-help-pro-se-litigants.

163.  *See Consumer Arbitration Fact Sheet*, AM. ARBIT. ASS'N, https://go.adr.org/consumer-arbitration (last visited Aug. 7, 2024) (the AAA provides a Small Claims Option under its Consumer Arbitration Rules. This option allows claimants to bring their claims in small claims court instead of arbitration. The AAA ensures that the process is accessible, affordable, and that arbitrators can grant relief available in court.); *see also* JAMS STREAMLINED ARB. RULES & PROCS. RULE 1 (JAMS 2021). JAMS offers a set of streamlined rules specifically for claims under $250,000. These rules emphasize efficiency, limiting discovery and focusing on resolving disputes through written submissions.

164.  However, it's important to note that doing so also pose several risks, such as inaccuracies (AI-generated documents may include incorrect or incomplete legal references), over-reliance (AI might not fully understand the nuances of a case or ethical implications), and lack of personalization (AI-generated documents may produce generic or poorly tailored arguments).

c. Arbitration as a Testing Ground for AI Integration
in the Judicial System

Deploying AI in arbitration offers a strategic starting point for integrating this technology within the broader judicial system. Arbitration has increasingly mirrored traditional court proceedings, with arbitrators determining their jurisdiction similarly to judges and the process involving extensive discovery phases much like those in litigation.[165] If AI can be successfully integrated into arbitration, it may set a precedent for its eventual adoption in the wider judicial context.

Currently, there's much debate about whether AI should be integrated into the judicial system. While AI can potentially improve access to justice by increasing the percentage of adequately represented litigants and reducing legal fees, critics argue that algorithmic tools lack transparency,[166] accountability,[167] and fairness.[168] In addition, there are concerns that AI's capability to handle complex legal reasoning is not yet sophisticated enough to navigate the nuances of law that human judges and lawyers can manage.[169] AI systems, particularly those trained on past legal decisions, may not adequately adapt to new legal standards or understand the context of human emotions and ethical considerations that are often crucial in legal judgments. This could result in decisions that are legally correct but morally or ethically questionable, potentially undermining public trust in the judicial system. Moreover, the use of AI could lead to a homogenization of legal outcomes.[170] As AI tools tend to generate solutions based on the most common interpretations of law, non-standard cases might not receive the individualized consideration they require. This could stifle the development of law, as precedents set by unique cases often lead to, or are reflections of significant legal reforms and social movements.

_______________

165. *See, e.g.*, Oliver Browne & Robert Price, *A Collision of Two Heads*, 82 Com. Litig. J. 18 (2018) (arbitration can be like domestic court litigations); *see also* Thomas J. Stipanowich, *Arbitration: The "New Litigation"*, 2010 U. Ill. L. Rev. 1 (2010).

166. *See, e.g.*, Danielle Keats Citron, *Technological Due Process*, 85 Wash. U. L. Rev. 1249, 1288–97 (2008); Natalie Ram, *Innovating Criminal Justice*, 112 Nw. U. L. Rev. 659 (2018); Rebecca Wexler, *Life, Liberty, and Trade Secrets: Intellectual Property in the Criminal Justice System*, 70 Stan. L. Rev. 1343 (2018).

167. *See, e.g.*, Margot E. Kaminski, *Binary Governance: Lessons from the GDPR's Approach to Algorithmic Accountability*, 92 S. Cal. L. Rev. 1529 (2019); Joshua A. Kroll et al., *Accountable Algorithms*, 165 U. Pa. L. Rev. 633 (2017); Anne L. Washington, *How To Argue with an Algorithm: Lessons from the COMPAS-ProPublica Debate*, 17 Colo. Tech. L.J. 131 (2018) (arguing for standards governing the information available about algorithms so that their accuracy and fairness can be properly assessed).

168. *See, e.g.*, Aziz Z. Huq, *Racial Equity in Algorithmic Criminal Justice*, 68 Duke L.J. 1043 (2019) (arguing that current constitutional doctrine is ill-suited to the task of evaluating algorithmic fairness and that current standards offered in the technology literature miss important policy concerns); Sandra G. Mayson, *Bias In, Bias Out*, 128 Yale L.J. 2218 (2019) (discussing how past and existing racial inequalities in crime and arrests mean that methods to predict criminal risk based on existing information will result in racial inequality).

169. *See* David Lat, *AI Use in Law Practice Needs Common Sense, Not More Court Rules*, Bloomberg L. (Feb. 28, 2024, 4:30 AM), https://news.bloomberglaw.com/us-law-week/ai-use-in-law-practice-needs-common-sense-not-more-court-rules.

As the number of trials declines, the rise of private arbitration offers an ideal testing ground for integrating AI. From 1962 to 2002, the percentage of federal civil cases resolved by trial decreased by 84%, with similar declines in state courts.[171] This shift may be due to concerns about litigation costs, delays, risks, and impacts on relationships. Lawsuits often strain personal and professional relationships, causing emotional distress and psychological consequences for plaintiffs and defendants.[172] The frequent adjournments, delays, and financial strain associated with litigation lead to many adverse emotional outcomes such as stress and sleepless nights.[173]

The decrease in civil litigation has spurred the growth of arbitration, offering perceived advantages like cost savings, shorter resolution times, expert decision-makers, privacy, and relative finality. [174] Arbitration's increasing judicialization, with

---

170. *See* Betsy Morris, *AI from AI: A Future of Generic and Biased Online Content?*, UCLA Anderson Rev. (Nov. 8, 2023), https://anderson-review.ucla.edu/ai-from-ai-a-future-of-generic-and-biased-online-content/.

171. Marc Galanter, *The Vanishing Trial: An Examination of Trials and Related Matters in Federal and State Courts*, 1 J. Empirical Legal Stud. 459, 460–63 (2004).

172. *See, e.g.*, Feikoab Parimah et al., *A Snapshot of Emotional Harms Caused by the Litigation Process – Qualitative Data from Ghana*, 2 Forensic Sci. Int'l: Mind & L. 100050 (2021); *Relationships During Litigation*, Physician Litig. Stress Res. Ctr., https://physicianlitigationstress.org/medical-malpractice-lawsuit-support-resources/relationships-during-litigation/ (last visited Aug. 7, 2024).

173. *Supra* note 172.

174. *See* Elena V. Helmer, *International Commercial Arbitration: Americanized, "Civilized," or Harmonized?*, 19 Ohio State J. on Disp. Resolution 35, 35–36 (2003) (discussing perceptions of the American influence on international arbitration); Amr A. Shalakany, *Arbitration and the Third World: A Plea for Reassessing Bias Under the Specter of Neoliberalism*, 41 Harv. Int'l L. J. 419, 434–35 arbitrators determining jurisdiction and pre-hearing procedures mirroring litigation, makes it a suitable environment to test AI integration.[175]

Another aspect in which arbitration mirrors litigation is the extensive pre-hearing procedures, such as discovery, including depositions.[176] While many arbitration rules and arbitrators strive to limit excessive discovery, [177] it is not uncommon for legal advocates to negotiate trial-like discovery procedures, sometimes adhering to standard civil procedural rules.[178] This tendency is exacerbated when arbitrators hesitate to enforce limits on these practices or adhere strictly to schedules, often

__________

(2000) (observing that international arbitration is no longer quicker than adjudication); *see also* Thomas J. Stipanowich, *ADR and the "Vanishing Trial": The Growth and Impact of "Alternative Dispute Resolution*,*"* 1 J. EMPIRICAL LEGAL STUD. 843, 895 (2004) (quoting Jeffrey Carr, Vice President & Gen. Counsel, FMC Tech., The Torch Is Passed, Corporate Counsel Panel, Remarks at the Annual Meeting of the CPR Institute for Dispute Resolution (Jan. 29–30, 2004)) (one corporate general counsel lamented, "we found arbitration generally is as expensive [as litigation] . . . less predictable, and not appealable. Arbitration is often unsatisfactory because litigators . . . run it exactly like a piece of litigation.").

175. *See* First Options of Chi., Inc. v. Kaplan, 514 U.S. 938, 947 (1995) (holding that absent contrary intent, the court is responsible for deciding arbitrability); JAMS COMPREHENSIVE ARB. RULES & PROCS. RULE 11 (JAMS 2021) (stating that once appointed, the arbitrator shall resolve disputes about the interpretation and applicability of these Rules and conduct of the arbitration hearing); *see also* Buckeye Check Cashing, Inc. v. Cardegna, 546 U.S. 440, 445-46 (2006) (enforcing a provision authorizing arbitrators to address arbitrability issues); Schlessinger v. Rosenfeld, Meyer & Susman, 47 Cal. Rptr. 2d 650, 659–60 (Cal. Ct. App. 1995) (concluding the arbitrator had implicit authority to rule on summary adjudication motions).

176. *See* Michael A. Doornweerd and Andrew F. Merrick, *Strategies for Controlling Discovery Costs in Commercial Arbitration*, 12 COM. & BUS. LIT. 4, 4 (2011) ("Arbitration hearings are often preceded by extensive discovery, including requests for voluminous document production and depositions.").

177. *See* Thomas J. Stipanowich et al., *Protocols for Expeditious, Cost-Effective Commercial Arbitration*, COLL. COM. ARBITRATORS (2010), https://nysba. org/NYSBA/Sections/Commercial%20Federal%20Litigation/ComFed%20 Display%20Tabs/Events/2019/Spring%20Meeting%20Materials/ Protocols%20for%20Expeditious,%20Cost-Effective%20Commercial%20 Arbitration.pdf.

178. Thomas J. Stipanowich, *Arbitration and Choice: Taking Charge of the "New Litigation" (Symposium Keynote Presentation)*, 7 DEPAUL BUS. & COM. L. J. 383, 393 n.36 (author saying that as an arbitrator, he has in the past encountered situations such that the counsel for arbitrating parties made a prior agreement to utilize the discovery provisions of the Federal Rules of Civil Procedure in arbitration).

because they do not want to alienate the parties and risk not being considered for future appointments.[179]

Historically, the U.S. has practiced the adoption of novel ideas in smaller or more contained environments before gaining wider acceptance. For example, the religious freedom experiments in the colonies, such as Rhode Island's separation of church and state and Pennsylvania's diverse religious freedoms, eventually became a cornerstone of the U.S. Constitution.[180] Similarly, in the judicial system, innovative uses of AI in arbitration could serve as a controlled experiment to test the technology's capabilities and address potential issues in a more contained and manageable environment. If these AI-driven methods prove effective and equitable in arbitration, they could lead to wider adoption across the broader judicial system, potentially transforming how justice is administered while ensuring that such technologies are introduced responsibly and ethically.

## II.
### The Critics are Killing the Baby

Despite the growing resistance to AI adoption in the legal domain, criticisms such as claims of AI being biased, discriminatory, lacking transparency, and accountability are insufficient grounds for outright rejection. The origins of discrimination and bias often lie within the human-provided data, not the AI mechanisms themselves. Given AI's early developmental stage and significant potential, maintaining an open environment that encourages its growth is essential. Adopting an overly moralistic tone could be counterproductive. The current regulatory frameworks, which typically involve supervising the system or focusing more carefully on human roles, may not be necessary for arbitration. As long as both parties agree by contract to this method of adjudication, it should be permitted.

---

179. *See* Clyde W. Summers, *Mandatory Arbitration: Privatizing Public Rights, Compelling the Unwilling to Arbitrate*, 6 U. Pa. J. Lab. & Emp. L. 685, 717 (2004) (arguing that arbitrators may be less restrictive with discovery than judges because of their concern over obtaining future appointment as an arbitrator).

180. *See* John M. Barry, *God, Government and Roger Williams' Big Idea*, Smithsonian Mag. (Jan 2012), https://www.smithsonianmag.com/history/god-government-and-roger-williams-big-idea-6291280/ (Williams persuaded Parliament to allow Rhode Island to divorce church and state, whose principle ultimately made its way into the Constitution); *see also* Pa. Const. §3.

## A.   *Resistance Against AI Does Not Offer Conclusive Reasons for Outright Rejection*

Criticizing AI has become trendy. Since the introduction of ChatGPT in December 2022, critics have proliferated in the literature. They argue that AI is biased and unfair, shaped by the initial data it receives.[181] When the underlying data is biased, the resulting algorithms can perpetuate discrimination and inequality.[182] For instance, AI can exhibit sexist behavior—defaulting to male doctors and female nurses in stories—because these patterns are reflected in the data it was trained on. [183] Critics also argue that AI displaces jobs, with the World Economic Forum estimating that AI could replace 85 million jobs by 2025 and more over time.[184] Concerns extend to privacy, as AI can predict psychological characteristics from digital footprints, infringing on individual privacy. [185] Additionally, AI's lack of transparency is another major concern, as leading companies do not share enough information about their foundation models' development and use.[186] Furthermore, AI's ever-increasing capabilities raise geopolitical concerns. [187]

Legal scholarship focusing on AI in arbitration echoes these criticisms. Many legal academics strongly discourage deploying AI in arbitration. [188] Professor David Horton contends that AI procedures do not qualify as "arbitration" under the FAA, which is predicated on human arbitrators.[189] Similarly, Professor Derick Lindquist and Ylli Dautaj argue that AI lacks human qualities like empathy, emotion, and life experience, which are crucial for justice and fairness.[190] Professor Lee-Ford Tritt notes that AI might enhance trust in arbitration by assisting human arbitrators but is less useful in complex, emotionally charged disputes where normative values are critical. [191] Gizem Halis Kasap highlights concerns about data integrity, machine biases, AI opacity, and lack of emotional intelligence, advising caution in adopting AI arbitrators. [192]

---

181. *See generally* Stephanie Bornstein, *Anti-discriminatory Algorithms*, 70 ALA. L. REV. 520, 522-523 (2019) (the decisions made by AI are shaped by the initial data it receives. If the underlying data is unfair, the resulting algorithms can perpetuate bias, incompleteness, or discrimination, creating potential for widespread inequality.).

182. *Id.*

183. *See* Haley Strack, *AI is Sexist, UN Women Claims*, NAT'L REV. (May 24, 2024), https://www.nationalreview.com/corner/ai-is-sexist-un-women-claims/.

184. *See* Mark Talmage-Rostron, *How Will Artificial Intelligence Affect Jobs 2024-2030*, NEXFORD UNIV. (Jan 10, 2024), https://www.nexford.edu/insights/how-will-ai-affect-jobs.

185. *See* S.C. Matz et al., *Psychological Targeting as an Effective Approach to Digital Mass Persuasion*, 114 PNAS 12714, 12714 (Nov. 13, 2017).

186. *See Transparency in AI Companies: Stanford Study,* LUMENOVA (Nov. 17, 2023), https://www.lumenova.ai/blog/how-transparent-are-ai-companies/.

187. *See* Barry Pavel et al., *AI and Geopolitics: How Might AI Affect the Rise and Fall of Nations?*, RAND (Nov. 3, 2023), https://www.rand.org/pubs/perspectives/PEA3034-1.html.

Another criticism focuses on datasets. Larger datasets do not necessarily improve AI performance; instead, they may perpetuate past patterns. Ian R. McKenzie et al.'s study observed that language models (LMs) do not always improve with increased size, despite common assumptions to the contrary.[193] This "inverse scaling" occurs because larger models may prefer repeating memorized sequences over adhering to specific contextual instructions.[194] They might replicate undesirable biases or focus on simpler subtasks within a complex setup, failing to address the main task effectively.[195]

Practical challenges also arise when applying AI in arbitration. General-purpose models like ChatGPT and Claude are not designed for specialized tasks such as large-scale international commercial arbitration.[196] Using AI in these contexts demands unprecedented trust from disputants. Before deployment, AI models must be fine-tuned to ensure confidentiality and evaluated for optimal trust-building methods. Ideally, these methods should reflect diversity in factors such as age, cultural background, gender, and socioeconomic status, including representation from both historically dominant and minority groups.

---

188. *Contra* Horst Eidenmueller & Faidon Varesis, What is an Arbitration? Artificial Intelligence and the Vanishing Human Arbitrator (July 15, 2020) (unpublished manuscript) (on file with SSRN) (arguing that "fully AI-powered arbitrations will be technically feasible and should be legally permissible at some point in the future. There's nothing in the concept of arbitration that requires human control, governance, or even input.").

189. *See* David Horton, *Forced Robot Arbitration*, 109 Cᴏʀɴᴇʟʟ L. Rᴇᴠ. 679, 721, 734 (2024).

190. *See* Derick H. Lindquist & Ylli Dauta, *AI in International Arbitration: Need for Human Touch*, 2021 J. Dɪsᴘ. Rᴇsᴏʟ. 39, 40-41 (2021); *see also* Cole Dorsey, *Hypothetical AI Arbitrators: A Deficiency in Empathy and Intuitive Decision-Making*, 13 Aʀʙ. L. Rᴇᴠ. 1, 31 (2021).

191. *See* Lee-Ford Tritt, *The Use of AI-Based Technologies in Arbitrating Trust Disputes*, 58 Wᴀᴋᴇ Fᴏʀᴇsᴛ L. Rᴇᴠ. 1203, 1223, 1243-44 (2023).

192. See Gizem Halis Kasap, *Can Artificial Intelligence Replace Human Arbitrators? Technological Concerns and Legal Implications*, 2021 J. Dɪsᴘ. Rᴇsᴏʟ. 209, 210 (2021).

193. *See* Ian R. McKenzie et al., Inverse Scaling: When Bigger Isn't Better (May 13, 2024) (unpublished manuscript) (on file with Transactions on Machine Learning Research).

194. *Id.*

195. *Id.*

But who are the programmers and policymakers to make decisions for the disputants? When the parties themselves believe that this technology is a suitable method to resolve their issues, why should others—academics, scientists, researchers—deny them this choice? This paternalistic culture should not prevail. Disputants, according to the framework in FAA, must have the autonomy to decide the means of resolving their conflicts, free from external imposition that may not fully understand or respect their specific needs and contexts.

The fact that AI exhibits bias is not sufficient grounds for its outright rejection in dispute resolution contexts. Ultimately, it is our behavior—"datafied" and scraped online—that forms the training datasets for these models. Why should we reject a tool that merely reflects ourselves?

It's fairly apparent from AI's working mechanism that we are the source of its problems. AI, as defined above, includes deep learning, LLMs, and ML. Each of these reflects the content of its training data. For instance, deep learning, a subset of ML, uses complex structures called neural networks to process data.[197] As input signals pass through these layers, each one performs calculations to predict an outcome.[198] These neural networks handle multiple layers of computations, similar to stacking several linear regression models interspersed with nonlinear functions to manage complex data relationships.[199] When an AI outputs something offensive, it is a reflection of the biases present in the training data, not an indication of inherent malice or error in the AI itself.

---

196. *See* Kasap, *supra* note 192 at 210.

197. *See What's the Difference Between Deep Learning, Machine Learning, and Artificial Intelligence?*, GOOGLE CLOUD, https://cloud.google.com/discover/deep-learning-vs-machine-learning?hl=en (last visited Aug 7, 2024).

198. *Id.*

199. *Id.*

LLM, a language model that functions by predicting the next word in a sequence, also reflects the biases and patterns present in its training data. Its working process involves inputting a sequence of words into the model, which then is used to predict what comes next.[200] The procedure uses a technique called self-supervised learning, which is like a student teaching themselves without explicit instructions.[201] Instead of being told what to learn, the model looks at the data it already has—like a text where the next word is the answer to a question it must guess. By predicting these next words repeatedly, the model teaches itself the patterns and rules of language, improving each time without needing a teacher to tell it if it's right or wrong.[202] The quality, diversity, and size of the dataset greatly influence the outcome — if the data contains inconsistencies, the model will learn these inaccuracies, leading to incorrect predictions. A non-diverse dataset will have limited capability to be generalized to unseen data and will generate results that favor certain language structures or content. Given its billions of parameters, it's possible that the model knows more about us than we do ourselves.

GPT (Generative Pre-Training Transformer), the basis of many pre-trained GenAI models, shares the same relationship for how the dataset affects the outcome. It uses an architecture optimized to focus on the most relevant parts of an input sequence at any given time.[203] This attention mechanism allows the model to efficiently process large amounts of data while prioritizing elements crucial for understanding context and generating responses.[204] The quality of the data determines the generated content in that the ideologies, perspectives, and assumptions embedded in the dataset are learned by the model.

In conclusion, rejecting AI due to its unsettling outputs is like rejecting a reflection of our societal flaws. Instead, we should keep an open mind and let AI grow, addressing the areas where improvement is needed.

---

200. *See* Andreas Stöffelbauer, *How Large Language Models Work – from Zero to ChatGPT*, MEDIUM (Oct. 24, 2023), https://medium.com/data-science-at-microsoft/how-large-language-models-work-91c362f5b78f.

201. *Id.*

202. *Id.*

203. *See What is GPT?*, AMAZON WEB SERVS., https://aws.amazon.com/what-is/gpt/ (last visited Aug. 7, 2024).

204. *Id.*

## B.  *Let AI Grow Under Favorable Conditions: Avoiding Overly Moralistic Views*

Social, cultural, and environmental norms, along with institutional requirements, often play a crucial role in the development and growth of technology. Focusing solely on AI's disadvantages and preventing parties from contracting for themselves can stifle technological growth and innovation. Overly moralistic views can prematurely stifle progress.

The development of economics illustrates the importance of welcoming external influences. Economics now holds a near-hegemonic status in strategic management science, not solely due to its internal rigor but also because of promotion by external entities like RAND and the Ford Foundation. During the 1930s Great Depression, classical economic theories failed. A new direction was essential. RAND's contributions, including funding and expertise, repositioned economics.[205] Pioneering economists like Kenneth Arrow and Harry Markowitz employed statistical methods, further promoting innovative approaches.[206]

Had the RAND Corporation and the Ford Foundation not prioritized the development of economics, the field might not have become as influential as it is today. Social science may have been delayed in using quantitative approaches to study complex phenomena; it's questionable whether the school of law and economics would have grown. It's highly probable that there might not be a systematic framework for a behavioral approach to economic analysis of law.

---

205.  *See A Brief History of RAND*, RAND, https://www.rand.org/about/history.html (last visited Aug. 7, 2024) (RAND's early contributions include developing theories and tools for decision-making under uncertainty. It also makes foundational contributions to game theory, linear and dynamic programming, mathematical modeling and simulation, network theory, and cost analysis – all critical theories and ideas in economics.).

206.  For Kenneth Arrow's contributions, see *Kenneth Arrow*, RAND, https://www.rand.org/pubs/authors/a/arrow_kenneth.html (last visited Aug. 7, 2024). For a biography of Kenneth Arrow and his position at RAND, see *Nobel Laureate Kenneth J. Arrow, a RAND Consultant Since 1948, Dies at 95*, RAND (Feb. 22, 2017), https://www.rand.org/news/press/2017/02/22.html. For Harry Markowitz's contribution to economic theory and his position at RAND, see Murray Coleman, *A Tribute to Harry Markowitz: In Memoriam of a Finance Legend*, INDEX FUND ADVISORS (June 28, 2023), https://www.ifa.com/articles/tribute-to-finance-legend-harry-markowitz.

Without the economic analysis of law, the field of legal scholarship would be profoundly different, as this perspective and analytical method have been applied across all domains of law—from torts to antitrust. For instance, in torts, behavioral economics has helped to predict and interpret how individuals might respond to various legal incentives and penalties.[207] In litigation strategies, game theory has been used to study parties' decision-making processes.[208] In antitrust, multi-layer modeling also plays a crucial role in the Department of Justice's evaluation regarding whether to allow certain acquisitions.[209] In criminal law, principles such as proportional punishment and marginal utility theory help ensure that fines and bail amounts are proportional to both the severity of the crime and the financial circumstances of the defendant.[210] Additionally, in health law, risk management and the supply and demand dynamics support the implementation of medical malpractice damage caps, aiming at curbing rising malpractice insurance costs and improving healthcare access and affordability.[211]

---

207. *See* Jay M. Feinman, *Incentives for Litigation or Settlement in Large Tort Cases: Responding to Insurance Company Intransigence*, 13 Roger Williams Univ. L. Rev. 189 (2008).

208. *See* Christopher Whitehouse & Simon Hart, *Game Theory and the Art of Litigation Strategy*, Reynolds Porter Chamberlain LLP (Apr. 2, 2019), https://www.rpc.co.uk/thinking/commercial-disputes/game-theory-and-the-art-of-litigation-strategy-article-4/; *see also* Engaging Experts, *How Game Theory Benefits Attorneys and Litigators*, Round Table Grp. (Jan. 15, 2021), https://www.roundtablegroup.com/expert-discussions/how-game-theory-benefits-attorneys-and-litigators/.

209. To investigate the merger's anti-competitive effects, DOJ used a variety of economics methods such as market concentration analysis (helping to predict if the merged entity could profitably raise prices; understanding consumer sensitivity to price changes in telecommunications services) and competitive effects analysis (assessing the impact of the merger on competitive dynamics, such as the potential for reduced innovation and the adverse effects on consumer choices and prices). Because the economic analysis predicted that the merger would significantly increase market concentration in the telecommunication sector, leading to higher prices, poorer service quality, and less consumer choice, DOJ filed a lawsuit in 2011 to block the merger; the FAA also signaled its opposition. In December 2011, AT&T and T-Mobile abandoned the merger. *See* Press Release, Office of Pub. Aff., U.S Dep't of Just., Justice Department Challenges AT&T/DirecTV's Acquisition of Time Warner (Nov. 20, 2017), https://www.justice.gov/opa/pr/justice-department-challenges-attdirectv-s-acquisition-time-warner. For an introduction of using economic analysis to predict the competitive effects of mergers, see Ken Heyer, *Predicting the Competitive Effects of Mergers by Listening to Customers*, Antitrust Div. U.S. Dep't of Just. (Sept. 2006), https://www.justice.gov/atr/predicting-competitive-effects-mergers-listening-customers.

210. *See* Richard A. Posner, *An Economic Theory of the Criminal Law*, 85 Colum. L. Rev. 1193, 1208, 1215–1216 (1985).

Indeed, if economics had faced the same level of criticism in the 1930s as AI does today, it simply might not have succeeded at all. The criticisms currently aimed at AI - relying on oversimplified models with unrealistic assumptions about human behavior, markets, and institutions; an overreliance on quantitative methods at the expense of qualitative insights; promoting market solutions that may disregard equity, justice, or sustainability; favoring policies that benefit wealthier segments of society while neglecting the needs of the poor and marginalized; applying methods to other fields like sociology, political science, and psychology without adequate consideration of their unique insights and methodologies; and exhibiting biases against gender and cultural diversity — are similar to those that have been directed at economics throughout its history, though they became more pronounced and diverse from the 1970s onwards.[212] Just as economics was allowed room to grow and

---

211. In 2003, Texas faced a significant crisis in medical malpractice insurance, which was driven by high damage awards and an increasing number of claims that led to high insurance premiums for healthcare providers. To evaluate and implement caps on medical malpractice damages, Texas charted the number of malpractice claims filed over time and the average size of the payment; it estimated the effects of various factors on malpractice insurance premiums and healthcare availability; it also gathered qualitative data from healthcare providers, insurance companies, and patients. *See Professional Liability Insurance Reform*, TEX. MED. ASS'N (Jan. 6, 2020), https://www.texmed. org/template.aspx?id=780; *see also* Patricia H. Born et al., *The Net Effects of Medical Malpractice Tort Reform on Health Insurance Losses: The Texas Experience*, 7 HEALTH ECON. REV. 42 (Nov 7, 2017), https://healtheconomicsreview. biomedcentral.com/articles/10.1186/s13561-017-0174-2.

212  For critiques about economics using oversimplified models, see Daniel Kahneman & Amos Tversky, *Prospect Theory: An Analysis of Decision Under Risk*, 47 ECONOMETRICA 263 (1979) (introducing Prospect Theory, challenging the classical assumption of rationality by illustrating how people actually make decisions under risk). For the same critique on AI, see Tim Pearce et al., Imitating Human Behaviour with Diffusion Model (Mar. 3, 2023) (unpublished manuscript) (on file with arXiv) (suggesting that using diffusion models to clone or replicate human behaviors has several limitations that cannot be overlooked: 1. Scalability - while diffusion models excel in capturing complex multimodal distributions, the computational power and scalability of these models can be challenging, especially in environments that require real-time interaction, 2. Sampling efficiency - diffusion model require iterative training, which is slow when generating actions in real-time scenarios; this affects the feasibility of applying these models in environments where decisions must evolve despite its imperfections, AI also deserves the opportunity to develop and mature without being prematurely judged. Just as economics needed support to achieve its potential, AI requires the same understanding and backing to realize its own transformative impact.

Another example is epidemiology. Initially, people did not believe in the causes of disease, attributing them to the ill wills of gods. However, the benefits of understanding the causes of diseases eventually led to widespread acceptance of epidemiological knowledge. For instance, in the mid-19th century, John Snow's work on the cholera outbreak in London faced significant skepticism. People believed in the miasma theory, thinking that bad air caused the disease. But Snow mapped the cholera cases around the Broad Street pump, which eventually led to the acceptance of his waterborne theory and laid the groundwork for modern epidemiology and public health practices.[213]

Currently, principles of epidemiology have been combined with legal analysis to understand the impact of laws on the health of populations. This field, known as legal epidemiology, examines how legal frameworks can be designed, implemented, and evaluated to improve public health.[214] For example, tobacco control laws have been informed by epidemiological evidence on the harms of smoking, leading to the implementation of various regulations such as advertising bans, smoking bans in

---

be made rapidly, 3. Dependent on quality and diversity of training data to generalize the computed results to human behaviors, 4. Handling sequential dependency - while the models can certainly handle sequential data, there remains a challenge in capturing long-term dependencies and ensuring that the sequence of actions generated is coherent over longer time horizons.). For the critique in economics about over-relying on quantitative methods at the expense of qualitative insights, see Edward Cartwright & Eghosa Igudia, *The Case for Mixed Methods Research: Embracing Qualitative Research to Understand the (Informal) Economy*, 2023 Rev. Dev. Econ. (Special Issue) 1 (Nov. 7, 2023) (saying that economics has long shunned qualitative research methods, and advocating for an integration of both). For critiques in economics about how it has neglected ethical considerations, see Hans J. Blommestein, *Why is Ethics Not Part of Modern Economics and Finance? A Historical Perspective*, Fin. & Common Good (2006), https://www.cairn.info/revue-finance-et-bien-commun-2006-1-page-54.htm.

213. *See* Fahema Begum, *Mapping Disease: John Snow and Cholera*, Royal Coll. of Surgeons of Eng. (Dec. 9, 2016), https://www.rcseng.ac.uk/library-and-publications/library/blog/mapping-disease-john-snow-and-cholera/.

214. *See Legal Epidemiology*, Ctrs. for Disease Control & Prevention (May 15, 2024), https://www.cdc.gov/cardiovascular-resources/php/toolkit/legal-epidemiology.html.

public places, and graphic warning labels on cigarette packs.[215] Similarly, laws requiring nutritional information on food packaging have been guided by epidemiological studies linking diet to chronic diseases like obesity, diabetes, and cardiovascular disease.[216] These regulations help consumers make healthier food choices.

Without allowing epidemiology the space to grow and be challenged, it wouldn't have achieved the impactful contributions it has made to public health today. The evolution of epidemiology underscores the importance of fostering emerging fields, providing them the opportunity to mature and demonstrate their potential, much like we must do with AI now.

Overly moralistic considerations can stifle innovations with potential global impact. Ensuring AI is fair is important, but when parties agree to use AI for dispute resolution, it is neither legally nor practically sensible for others to contest this usage. Constantly adopting a moralistic stance can hinder progress. An example is Cardinal Richelieu, who served as the First Minister of France from 1624 to 1642. Willing to break away from the prevailing ideologies of his time, he transformed France into a leading power of the 17th century.

When Richelieu took office, the Holy Roman Emperor Ferdinand II was working to restore Catholic dominance and suppress Protestantism,[217] aiming to unite Christian Europe under the Emperor and the Pope. However, the Empire—spanning Germany and Northern Italy—never fully achieved this "universality" due to tensions between the Emperor's political ambitions and the Church's spiritual authority. These clashes often arose as secular rulers, including the Habsburgs, sought regional autonomy by challenging the Church.

Habsburg expansion threatened France, which was encircled by Habsburg territories: Spain to the south, Spanish-influenced Italian states to the southeast, and Franche-Comté

---

and the Spanish Netherlands to the north.[218] Richelieu, though a Catholic cardinal, countered this by aligning with Protestant princes, issuing the Grace of Alais in 1629 to grant French Protestants religious freedom.[219] He also supported Protestant German princes and subsidized Gustavus Adolphus of Sweden to reduce Habsburg threats, prioritizing France's security over religious lines.[220]

By focusing squarely on preventing the emergence of a major power on France's borders, Richelieu consciously steered clear of ideological debates. Had he succumbed to the religious zeal and ideological fanaticism prevalent in his era, France might not have been able to maintain its distance while Germany was being devastated. If constrained by his own morality, the French monarchy might not have risen to the formidable power it later became. Richelieu's Machiavellian pragmatism provided the intellectual foundation for pursuing foreign policies that prioritized state interests over moral or religious considerations.[221] This approach not only fortified France but also laid the groundwork for the later development of political realism, shaping the conduct of states in subsequent centuries to come.[222] Granted, this theory has been criticized widely for its lack of moral foundation and contribution to power politics. Nevertheless, it's undeniable that this doctrine played a crucial role in forestalling German unification by some two centuries.

In the context of AI in arbitration, adopting a pragmatic approach can lead to substantial advancements. The underlying principle here is intellectual humility—we should recognize that it is difficult to predict what will work and remain open to experimentation. Utilizing AI for dispute resolution in contractual agreements allows for innovative approaches that might streamline processes and yield fair outcomes efficiently. By eschewing rigid moralistic frameworks and embracing a trial-and-error methodology, we can uncover new pathways to solving complex problems.

---

218. *See* Henry Kissinger, Diplomacy 59 (Simon & Schuster, 1994).

219. *Id.* at 61–62.

220. *Id.*

221. *Id.* at 63–67 (discussing the impact of Richelius's pragmatic approach to national interests in diplomacy).

222. *Id.*

## C. *Arbitration Should Allow Flexible, Contract-Based Experimentation in a Fast-Evolving Regulatory Landscape*

Given the fluid landscape of AI regulation, arbitration should provide a space for experimentation based on contractual agreements between parties. The two primary current approaches to AI regulation are unnecessary when parties can agree by contract on their preferred method of dispute resolution. As a developing technology, AI should be allowed room for innovation, with a hands-off approach fostering creativity and letting the market shape AI integration and application.

The first regulatory approach focuses on the specific use of AI. This branch of regulation incorporates a strategy that involves adding a human element to the decision-making process.[223] It aims to address generalized social injustice, accountability, oversight, and algorithm auditing. However, it fails to specify why a human is or isn't involved or clarify the roles the human is supposed to play. Additionally, it doesn't account for the human's needs, skills, or weaknesses.[224] These regulations are based on the unexamined assumption that humans can effectively oversee algorithmic decision-making, despite the often-flawed design of the interfaces.[225] The underlying belief is that because humans are tangible and identifiable, they can be directly regulated as familiar targets.[226]

More specifically, these regulations focus on various aspects of AI deployment. For instance, some address implementation questions, advocating for procedural regularity and oversight to ensure fairness and accuracy in AI scoring systems.[227] Others consider the potential social effects, such as the dispropor-tionate impact of AI on socially marginalized groups.[228] Some evaluate related social and governance considerations, sug-gesting that states and industry should collaborate to govern the use of algorithms and public data to prevent private AI governance.[229] Other works examine how humans influence algorithmic decision-making in specific tasks.[230]

---

223. *See, e.g.*, Meg Leta Jones, *The Right to a Human in the Loop: Political Constructions of Computer Automation and Personhood*, 47 Soc. Stud. Sci. 216, 224 (2017) (describing the EU's use of the human in the loop as a regulatory tool).

224. *See* Rebecca Crootof, Margot E. Kaminski & W. Nicholson Price II, *Humans in the Loop*, 76 Vand. L. Rev. 429, 437 (2023).

225. *See, e.g.*, Ben Green, *The Flaws of Policies Requiring Human Oversight of Government Algorithms*, 45 Comput. L. & Sec. Rev. 1 (2022).

226. *See, e.g.*, Ashley Deeks, Noam Lubell & Daragh Murray, *Machine Learning, Artificial Intelligence, and the Use of Force by States*, 10 Nat'l Sec. L. & Pol'y 1, 4 (2019) (predicting that states will use machine learning to aid decisions about whether to use force against or inside another state); Ashley S. Deeks, *Predicting Enemies*, 104 Va. L. Rev. 1529, 1530–31 (2018) (noting that leaders are encouraging the use of machine learning to improve military capabilities and decision-making).

227. *E.g.*, Danielle Keats Citron & Frank A. Pasquale, *The Scored Society: Due Process for Automated Predictions*, 89 Wash. L. Rev. 1, 19 (2014).

Beyond specific uses, these policies are also proposed within their respective fields. For example, international humanitarian law mandates that military use of AI must comply with applicable international laws.[231] Health law requires that AI applications in healthcare prioritize patient safety, maintain confidentiality, and adhere to existing medical standards and regulations.[232] Administrative law must adapt to the reality of automation, using technology to enhance the foundational

---

228. *See, e.g.*, Rebecca Crootof, "*Cyborg Justice" and the Risk of Technological–Legal Lock-In*, 119 COLUM. L. REV. F. 233, 235 (2019) (claiming that convert-ing legal procedures and judicial decision-making into code could introduce new obstacles to legal development, potentially leading to stagnation and undermining the legitimacy of the judiciary); Richard M. Re & Alicia Solow-Niederman, *Developing Artificially Intelligent Justice*, 22 STAN. TECH. L. REV. 242, 247 (2019) (claiming that the incorporation of AI into the common law judi-cial system will alter expectations about the judiciary, shifting focus from pro-cedural fairness to a greater emphasis on outcomes).

229. Alicia Solow-Niederman, *Administering Artificial Intelligence*, 93 S. CAL. L. REV. 633, 690 (2019); Margot E. Kaminski, *Binary Governance: Lessons from the GDPR's Approach to Algorithmic Accountability*, 92 S. CAL. L. REV. 1529, 1538 (2019) (arguing that algorithmic decision making relies on flawed human decisions as well. States and industry ought to govern together).

230. *See, e.g.*, Kiel Brennan-Marquez, Karen Levy & Daniel Susser, *Strange Loops: Apparent Versus Actual Human Involvement in Automated Decision Making*, 24 BERKELEY TECH. L.J. 745, 749 (2019) (discussing how having a human in the loop affects the decision-making process); Meg Leta Jones, *The Ironies of Automation Law: Tying Policy Knots with Fair Automation Practices Principles*, 18 VAND. J. ENT. & TECH. L. 77, 90–91 (2015) (describing how people automate the easiest tasks to oversee fallible automated systems, while simultaneously increasing the amount of time they use to resolve the more difficult issues).

231. *See Political Declaration on Responsible Military Use of Artificial Intelligence and Autonomy*, U.S. DEP'T OF STATE (Nov. 9, 2023), https://www.state.gov/political-declaration-on-responsible-military-use-of-artificial-intelligence-and-autonomy-2/.

232. *See, e.g.*, W. Nicholson Price II, *Regulating Black-Box Medicine*, 116 MICH. L. REV. 421, 423–24 (2017) (suggesting a more collaborative approach to gov-ern medical algorithm); *see also* Ciro Mennella et al., *Ethical and Regulatory Challenges of AI Technologies in Healthcare: A Narrative Review*, 10 HELIYON 1, 9 (2024).

purposes of the administrative state, such as expertise, agility, and discretion in governance.[233]

One significant drawback is the vagueness regarding how users should use algorithms to reach decisions. For example, a Colorado law governing the use of facial recognition by government agencies requires human involvement when making decisions that produce legal effects concerning individuals.[234] While the law mandates that decisions be "subject to meaningful human review," defined as "review or oversight by one or more individuals who are trained [in accordance with the statute's requirements] and have the authority to alter a decision under review,"[235] it does not specify what the training should entail, nor does it clearly define the scope of "legal effects concerning individuals." There is no data that suggests an added layer of human review makes the system better. In today's world, any effect relating to humans could be interpreted as legal if viewed broadly. Furthermore, the law does not detail the procedural aspects of human review, such as whether the review is automatic for every decision, how many reviewers are required, or how disputes among reviewers should be resolved.

A second, more holistic approach recognizes that the system is more than its parts.[236] Since humans are involved in every phase of the algorithm, from training to deployment, a broader definition of the system's tasks is crucial.[237] This approach identifies nine distinct categories of human-AI interactions, including corrective, resilience, justificatory, dignitary, accountability, stand-in, friction, warm body, and interface roles.[238]

---

233. *See e.g.*, Ryan Calo & Danielle Keats Citron, *The Automated Administrative State: A Crisis of Legitimacy*, 70 Emory L.J. 797, 800 (2021) (noting the trend in state and federal public benefits agencies towards incorporating automated systems).

234. Colo. Rev. Stat. § 24-18-303 (2022); *see generally* S.B. 22-113, 73d Gen. Assemb., Reg. Sess. (Colo. 2022).

235. Colo. Rev. Stat. §§ 24-18-303, -301(III)(9) (2022).

236. *See* Crootof, Kaminski & Price, *supra* note 224, at 437.

237. *Id.* at 444.

238. For a more thorough explanation of the nine categories, see *id.* at 473–487 (corrective Roles, where humans enhance system performance by correcting errors; Resilience Roles, acting as fail-safes to stop the system during emergencies and prevent malfunctions; and Justificatory Roles, where humans provide reasoning for decisions, increasing the system's legitimacy. Dignitary Roles protect the dignity of individuals affected by system decisions, while Accountability Roles involve humans in allocating liability and administering censure when necessary. Stand-in Roles see humans representing or substituting for other humans and human values, ensuring alignment with However, cataloging these roles exhaustively is impossible, as technology reveals new applications and roles for humans that were previously unimaginable. For instance, when ChatGPT launched in 2023, its potential to act as a realistic, on-demand emotional companion was not immediately recognized. Yet by 2024, it had captured media attention for its ability to simulate intimacy.[239] Similarly, by March of that year, Suno was creating high-quality music from text prompts, and AlphaFold3 was accurately predicting biomolecular structures—developments that had not been anticipated.[240] All these advancements raise crucial legal issues that previously were not thought of. It is a futile attempt to create a comprehensive list of roles that need regulation.

As a result, one might argue for adopting industry standards for regulation, based on the premise that developers of the technology best understand how to use it safely.[241] For instance, if computer scientists believe that AI should not replace humans as arbitrators, then these systems should be rigorously regulated for safety and fairness before deployment. However, AI technology is still in development, and universally accepted best practices have not been established. Unlike biochemistry, which benefits from decades of peer review and established practices, AI lacks a comprehensive body of long-term data to set standards. Technologies like deep learning have only been in widespread use for the past decade, previously limited by

---

societal norms. Friction Roles involve slowing down automated decision-making to allow for oversight and reflection. Additionally, Warm Body Roles help preserve human jobs by maintaining a human presence in automated processes, and Interface Roles serve as the critical link between the system and its users, facilitating interaction and understanding.).

239. *See* Kevin Roose, *Meet My A.I. Friends*, N.Y. TIMES (May 9, 2024), https://www.nytimes.com/2024/05/09/technology/meet-my-ai-friends.html (showing that LLM can now serve as emotional companions for people).

240. *See* Brian Hiatt, *A ChatGPT for Music is Here. Inside Suno, the Startup Changing Everything*, ROLLING STONE (Mar. 17, 2024), https://www.rollingstone.com/music/music-features/suno-ai-chatgpt-for-music-1234982307/; *see also* Google DeepMind AlphaFold Team & Isomorphic Labs, *AlphaFold 3 Predicts the Structure and Interactions of All of Life's Molecules*, GOOGLE (May 8, 2024), https://blog.google/technology/ai/google-deepmind-isomorphic-alphafold-3-ai-model/.

241. *See, e.g.*, Nicholas Bagley & Richard L. Revesz, *Centralized Oversight of the Regulatory State*, 106 COLUM. L. REV. 1260 (2006) (arguing for reconsidering the foundational role of centralized review in regulation, and supporting the idea of adopting industry standards for regulations, based on the premise that those who develop technology often best understand its safe and effective use).

data availability and computing power. Furthermore, emerging applications such as autonomous vehicles and predictive policing are still evolving, without extensive, validated research for guidance. Relying solely on industry standards may overlook the diverse ways in which users' backgrounds, personalities, and environmental conditions affect their interactions with AI. Therefore, AI regulation requires a flexible and adaptive framework that can evolve with technological advancements, ensuring these regulations remain relevant and effective in addressing the challenges of AI.

Arbitration, as a consensual means of resolving disputes, could bypass certain regulations. This flexibility allows parties to freely experiment, providing opportunities to observe and study AI's effectiveness and potential biases. Arbitration agreements can include specific provisions for the use of AI, ensuring that both parties are aware of and agree to the use of AI in the dispute resolution process. Of course, we know that these agreements will not directly mimic the outcome that might be reached by a court, and that is exactly what ADR is designed to permit.

Arbitration also provides a unique opportunity for AI experimentation because it operates outside the traditional court system. Parties can tailor arbitration procedures to their specific needs, including the integration of AI. This flexibility allows for the development and testing of AI technologies in a controlled environment, providing valuable insights into their strengths and weaknesses.

In conclusion, critics' concerns about AI's biases, discrimination, and lack of transparency are insufficient for its outright rejection. The origins of these issues lie within the human-provided data, not the AI mechanisms themselves. By allowing AI to grow under favorable conditions and adopting a flexible regulatory framework, we can harness AI's potential while addressing its challenges responsibly. Arbitration, with its inherent flexibility, provides an ideal testing ground for AI integration in the broader judicial system.

## III.

### Future Direction and Conclusion

Arbitration presents an unparalleled opportunity to lead the AI revolution in the legal domain. It is a field inherently designed to empower parties to define their own terms of justice through mutual agreements. Integrating AI into arbitration honors this principle, providing a platform for innovative adjudication methods that are both cost-effective and efficient.

In a climate rife with AI skepticism, it is crucial to maintain the integrity of arbitration agreements. The right of parties to choose AI as their arbitrator must be respected, ensuring that arbitration remains a flexible and dynamic alternative to traditional litigation. This respect for contractual autonomy not only upholds the fundamentals of arbitration law but also paves the way for the evolution of AI jurisprudence, a critical advancement for our legal system.

Drawing from the profound insights of Judge Richard Posner, we adapt his thoughts to advocate fervently for AI in alternative dispute resolution. The ethical obligations of AI arbitrators can only be comprehended by acknowledging the fundamental differences between AI adjudication and traditional human adjudication. Consent to arbitrate a commercial dispute with AI is entirely voluntary, rooted in the contractual agreement. This voluntariness is a pivotal safeguard absent in the coercive nature of the judicial system. The historical American fear of government oppression has fostered a judiciary where impartiality is prioritized over specialized expertise.[242] Conversely, those who opt for AI arbitration do so because they value a tribunal endowed with unparalleled data processing capabilities and consistency as well as reduced costs, far exceeding the limited scope and high cost of generalist courts. The technological prowess of AI arbitrators is especially beneficial in commercial disputes, characterized by their complexity and the vast datasets involved.

It is essential to avoid prematurely dismissing the potential of AI in ADR. Instead, we must nurture this innovative approach, allowing it to mature and demonstrate its full capabilities. By fostering an environment conducive to the growth of AI arbitration, we stand to achieve significant advancements in legal adjudication, ensuring that this groundbreaking technology is given a fair and unbiased chance to prove its worth.

In summary, let us not stifle the promising AI-ADR initiative. Rather, let us support its growth, carefully observe its development, and assess its capabilities as it evolves. By doing so, we can unlock a future where AI enhances the efficiency and fairness of arbitration, benefiting all parties involved.

---

242. Merit Ins. Co. v. Leatherby Ins. Co., 714 F.2d 673, 679 (7th Cir. 1983).